%% file: main.tex
\documentclass[10pt,twocolumn,letterpaper]{article}

\usepackage{cvpr}    
\usepackage{amsmath}   
\usepackage{amssymb}   
\usepackage{mathtools} 
\usepackage{bm}        
\usepackage{graphicx}  
\usepackage{epstopdf}
\usepackage[pagebackref,breaklinks,colorlinks]{hyperref} 
\usepackage{colortbl}
\usepackage[utf8]{inputenc} 
\usepackage{booktabs,tabularx,array,makecell}
\input{preamble}

\def\paperID{7646} 
\def\confName{CVPR}
\def\confYear{2026}
\vspace{-3mm}
\title{Percept-WAM: Perception-Enhanced World-Awareness-Action Model for Robust End-to-End Autonomous Driving}

\author{
Jianhua Han\textsuperscript{1}\thanks{Equal contribution.} \quad
Meng Tian\textsuperscript{1}\footnotemark[1] \quad
Jiangtong Zhu\textsuperscript{1}\footnotemark[1] \quad
Fan He\textsuperscript{1} \quad
Huixin Zhang\textsuperscript{1} \quad
Sitong Guo\textsuperscript{1} \quad
Dechang Zhu\textsuperscript{1} \\
Hao Tang\textsuperscript{1} \quad
Pei Xu\textsuperscript{1} \quad
Yuze Guo\textsuperscript{1} \quad
Minzhe Niu\textsuperscript{1} \quad
Haojie Zhu\textsuperscript{1} \quad
Qichao Dong\textsuperscript{1} \quad
Xuechao Yan\textsuperscript{1} \\
Siyuan Dong\textsuperscript{1} \quad
Lu Hou\textsuperscript{1} \quad
Qingqiu Huang\textsuperscript{1} \quad
Xiaosong Jia\textsuperscript{2} \quad
Hang Xu\textsuperscript{1}\thanks{Corresponding author.} \\
\textsuperscript{1}Yinwang Intelligent Technology Co. Ltd., \textsuperscript{2}Fudan University
}

\begin{document}
\maketitle
\thispagestyle{plain} 
\input{sec/0_abstract}    
\input{sec/1_intro}

\input{sec/2_related_work}
\input{sec/3_methods}
\input{sec/4_experiments}
\input{sec/5_conclusion}
\newpage
{
    \small
    \bibliographystyle{unsrt}
    \bibliography{main}
}

\appendix
\input{sec/X_suppl}

\newpage

\end{document}

%% file: preamble.tex
\usepackage{multirow}

%% file: sec/0_abstract.tex
\vspace{-2mm}
\begin{abstract}
Autonomous driving heavily relies on accurate and robust spatial perception. 
Many failures arise from inaccuracies and instability, especially in long-tail scenarios and complex interactions. 
However, current vision–language models are weak at spatial grounding and understanding, and VLA systems built on them therefore show limited perception and localization ability.
To address these challenges, we introduce Percept-WAM, a perception-enhanced World-Awareness-Action Model that is the first to implicitly integrate 2D/3D scene understanding abilities within a single vision-language model (VLM).
Instead of relying on QA-style spatial reasoning, Percept-WAM unifies 2D/3D perception tasks into World-PV and World-BEV tokens, which encode both spatial coordinates and confidence.
We propose a grid-conditioned prediction mechanism for dense object perception, incorporating IoU-aware scoring and parallel autoregressive decoding, improving stability in long-tail, far-range, and small-object scenarios. 
Additionally, Percept-WAM leverages pretrained VLM parameters to retain general intelligence (e.g., logical reasoning) and can output perception results and trajectory control outputs directly. 
Experiments show that Percept-WAM matches or surpasses classical detectors and segmenters on downstream perception benchmarks, achieving 51.7/58.9 mAP on COCO 2D detection and nuScenes BEV 3D detection. When integrated with trajectory decoders, it further improves planning performance on nuScenes and NAVSIM, 
e.g., surpassing DiffusionDrive by 2.1 in PMDS on NAVSIM.
Qualitative results further highlight its strong open-vocabulary and long-tail generalization.
\end{abstract}
\vspace{-2mm}

%% file: sec/1_intro.tex
\vspace{-4mm}
\section{Introduction}
\label{sec:intro}
\vspace{-1mm}
\begin{figure}[t!]
  \vspace{-3mm}
  \centering
  \includegraphics[width=\linewidth]{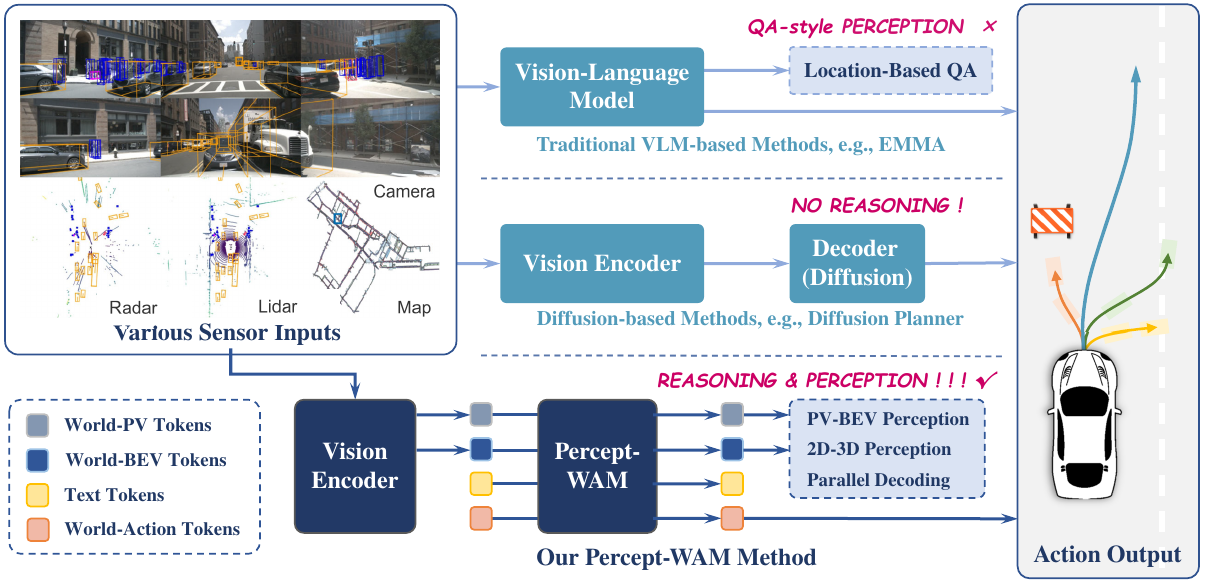}
  \vspace{-5mm}
  \caption{
  \textbf{Comparison of VLA paradigms.} (i) QA-style supervision frames spatial understanding as question answering \cite{hwang2024emma}, providing only indirect localization and yielding weak perception. (ii) Diffusion–decoder pipelines \cite{zheng2025diffusion} offer generative control but lack LLM-level reasoning. (iii) Percep-WAM implicitly integrates 2D/3D scene understanding within a single VLM, enabling robust perception and trajectory prediction in complex scenarios.
}
  \label{fig: Introduction}
  \vspace{-4mm}
\end{figure}

Autonomous driving ultimately relies on precise spatial perception and the capability to reason about the environment to make appropriate control decisions \cite{fan2023autonomous,tian2025nuscenes,guo2024drivemllm}.
In practice, small geometric errors (e.g., detection bias, yaw drift, or BEV/occupancy mistakes) snowball into brittle decisions, especially under long-tail conditions (e.g., night, rain, and small or rare objects).
Meanwhile, many recent VLA frameworks~\cite{wang2025alpamayo,fu2025orion,xu2024vlm,hwang2024emma,wen2025dexvla,wang2025rad,pan2025omnimanip,argus2025cvla} leverage VLM backbones  to incorporate reasoning into driving control in complex scenarios. However, recent evaluations and benchmarks~\cite{yang2025thinking,daxberger2025mm,cheng2024spatialrgpt} show that general-purpose VLMs still struggle in core spatial abilities (3D localization drift, temporal inconsistency and unreliable confidence), indicating that broad vision–language alignment does not guarantee geometric competence. This gap directly impacts closed-loop performance and stability on real-world routes and scenario-rich benchmarks~\cite{jia2024bench2drive,dauner2024navsim,caesar2020nuscenes}.

Many existing methods \cite{hwang2024emma,zeng2025futuresightdrive,chi2025impromptu,wang2024drivecot} formulate spatial understanding as QA-style supervision (in \Cref{fig: Introduction}), e.g., asking “What is the distance to the moving object ahead?”. However, this provides only indirect localization signals, rarely yields persistent, localizable world states, and leads to duplicate detections with poorly calibrated confidence in crowded scenes~\cite{wang2023visionllm,driess2023palm}. 
On the other hand, some methods \cite{zheng2025diffusion,liao2025diffusiondrive,zhao2025diffe2e,liu2025bridgedrive} abandon LLM-based architectures and adopt an encoder–diffusion-decoder pipeline to perform end-to-end (E2E) trajectory prediction directly. However, omitting explicit spatial task learning degrades E2E performance, and complex autonomous-driving scenarios also require the reasoning capacity of LLMs~\cite{mao2023gpt,wang2024drivecot,zhang2024think}.
These limitations motivate embedding explicit, persistent world states within a single VLM and jointly optimizing perception and trajectory to improve robustness—conceptually aligned with planning-oriented frameworks such as UniAD~\cite{hu2023planning}, but instantiated within a VLM.

Therefore, we propose the \textbf{Percept-WAM} method, a World–Awareness–Action framework that embeds \emph{world states} within a single VLM. Percept-WAM unifies 2D and 3D perception via two token families, \textbf{World-PV} (image-plane) and \textbf{World-BEV} (bird’s-eye view), each encoding metric coordinates together with calibrated confidence, providing localized, reusable evidence for downstream reasoning and planning. 
Specifically, Percept-WAM initializes from a pretrained VLM to preserve general capabilities (e.g., logical reasoning), and includes (i) a grid-conditioned prediction head that structures dense multi-object inference, (ii) an IoU-aware scoring objective that explicitly calibrates detection confidence, and (iii) parallel autoregressive decoding that maintains throughput without sacrificing stability.
Furthermore, we introduce the \textbf{World-Action} tokens to align and fuse perceptual information, and a lightweight MLP decoder for accurate and efficient future-trajectory prediction.
In summary, within a single backbone, the model can either output both perception results (e.g., 2D/3D bounding boxes) and trajectory, or only trajectory.
We also equip Percept-WAM with efficient streaming inference technique to meet the demand of real-world applications. 

In experiments, Percept-WAM matches or surpasses strong detector and segmenter baselines on downstream perception benchmarks, achieving 51.7 mAP on COCO~\cite{lin2014microsoft} 2D detection and 58.9 mAP on nuScenes~\cite{caesar2020nuscenes} BEV 3D detection, surpassing the 47.5/52.3 mAP of LMM-Det~\cite{li2025lmm} and PointPillars~\cite{lang2019pointpillars}, respectively. 
Planning performance on nuScenes and NAVSIM~\cite{dauner2024navsim} is further improved by equipping a query-based action decoder. It obtains 90.2 PMDS on NAVSIM, outperforming  DiffusionDrive~\cite{liao2025diffusiondrive} by 2.1, and exhibits a short frame latency of 707 ms via streaming inference. 
Qualitative evidence further shows robust open-vocabulary and long-tail generalization, underscoring enhanced perception-to-action capability within a unified World-Awareness-Action model.

We summarize our contributions as follows:
\begin{itemize}   
    \item \textbf{Perception-enhanced World Tokens.} Percept-WAM builds the first framework to seamlessly unify 2D/3D perception within a single VLM via \emph{World-PV} and \emph{World-BEV} tokens, which encode metric coordinates and calibrated confidence, yielding reusable, localizable world states for reasoning and control. 
    \item \textbf{Grid-conditioned Dense Perception.} We introduce a novel grid-conditioned head, augmented with \emph{IoU-aware} scoring and \emph{parallel AR} decoding. This design significantly improves both accuracy and stability in long-tail, far-range, and small-objects perception. 
    \item \textbf{Perception-to-Action Paradigm.} 
    Beyond outperforming or matching SOTA detector/segmenter baselines on \emph{nuImages} and \emph{nuScenes}, Percept-WAM shows outstanding planning performance on \emph{nuScenes} and \emph{NAVSIM}, through the alignment of World-PV, World-BEV and World-Action tokens.
\end{itemize}

%% file: sec/2_related_work.tex
\section{Related Work}

\begin{figure*}[t!]
  \vspace{-2mm}
  \centering
  \includegraphics[width=0.85\linewidth]{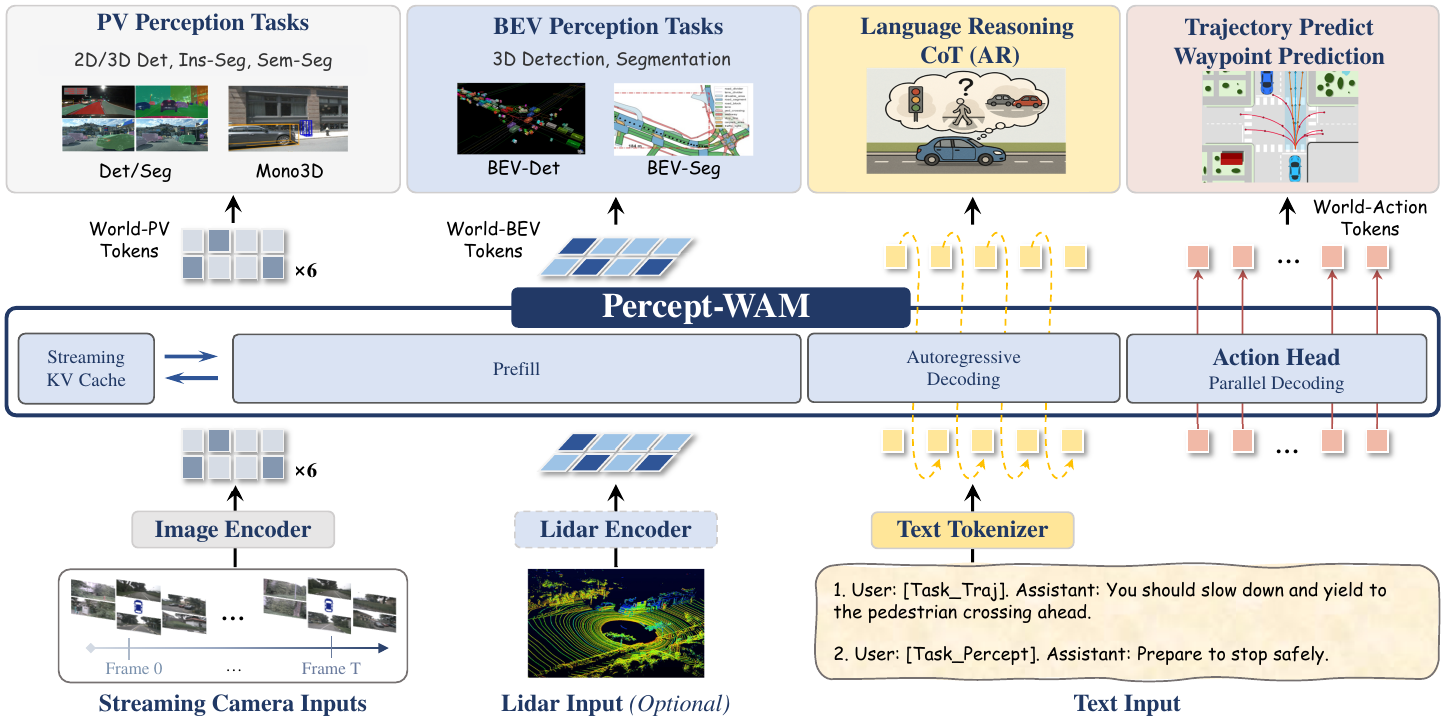}
    \vspace{-2mm}
  \caption{\textbf{The overall architecture of Percept-WAM}. 
  i) We use a \textbf{pretrained VLM backbone} to maintain general reasoning capability, ii) Percept-WAM unifies 2D and 3D perception via \textbf{World-PV and World-BEV tokens}. The learnable BEV-Level grid tokens implicitly model the mapping from PV features to BEV-space representations. 
  iii) \textbf{An Action Head} is introduced to predict trajectories from world tokens via parallel decoding.
  An additional memory bank is introduced to support efficient streaming inference.}
  \label{fig: architecture}
  \vspace{-3mm}
\end{figure*}

\noindent\textbf{VLMs/VLAs and Spatial Grounding.}
Early foundation VLMs (e.g., CLIP~\cite{radford2021learning}) enable broad vision–language alignment and zero-shot transfer, and subsequent instruction-tuned architectures (Flamingo~\cite{alayrac2022flamingo}, BLIP-2~\cite{li2023blip}, LLaVA~\cite{liu2023visual}) improve few-shot adaptation with interleaved inputs. 
Yet recent studies report persistent gaps in metric geometry, relative spatial relations, and BEV/top-view reasoning~\cite{cheng2024spatialrgpt,stogiannidis2025mind}. 
In parallel, Vision–Language–Action (VLA) models demonstrate that injecting action supervision into pretrained VLMs boosts embodied control (RT-2~\cite{brohan2024rt}, OpenVLA~\cite{kim2024openvla}), and driving-oriented VLM/VLA systems explore language-conditioned planning (DriveVLM~\cite{tian2024drivevlm}, DriveLM~\cite{sima2024drivelm}). 
These findings collectively motivate \emph{encoding explicit geometry} rather than relying  on QA-style supervision. Our Percept-WAM instantiates this by representing world states allowing the backbone to reason over spatially grounded evidence and to reuse tokens for downstream tasks.

\noindent\textbf{BEV Perception and 3D Occupancy.}
BEV representations have become a de-facto interface for 3D understanding in AD: Lift–Splat–Shoot (LSS)~\cite{philion2020lift} lifts features to frustums and rasterizes them into BEV grids; BEVDepth~\cite{li2023bevdepth} improves depth quality with explicit supervision; and BEVSegFormer~\cite{peng2023bevsegformer} tailors deformable attention for BEV semantic segmentation. 
Beyond planar BEV, OccFormer~\cite{zhang2023occformer} pushes toward 3D semantic occupancy, enriching vertical structure and long-range context. 
For multi-sensor fidelity, BEVFusion~\cite{liu2022bevfusion} disentangles camera–LiDAR fusion so the camera stream remains predictive under LiDAR dropouts. 
Percept-WAM packs spatial evidence as reconstructable World BEV tokens and parallel autoregressive decoding for 3D perception.

\noindent\textbf{E2E AD Model and Closed-Loop Evaluation.}
Jointly optimizing perception and planning reduces error compounding. TransFuser~\cite{chitta2022transfuser} integrates multi-modal attention for closed-loop control, while UniAD~\cite{hu2023planning} exemplifies planning-oriented full-stack learning. 
Surveys further codify challenges such as interpretability, world modeling, and causal confusion in E2E AD~\cite{chen2024end}. 
Beyond open-loop L2 metrics, recent benchmarks move to scenario-driven closed-loop testing: Bench2Drive~\cite{jia2024bench2drive} evaluates multi-ability behaviors and nuPlan~\cite{caesar2021nuplan} establishes large-scale real-world planning protocols; NAVSIM~\cite{dauner2024navsim} enables data-driven pseudo-simulation at scale. 

\noindent\textbf{Reasoning VLM/VLA for Driving.}
Building on driving-centric VQA and language datasets, recent works focus on \emph{reasoning} that spans perception, prediction, and planning~\cite{qian2024nuscenes}. WOMD-Reasoning~\cite{li2024womd} scales to millions of Q\&As on interactive behaviors in the Waymo Open Motion Dataset. DRAMA~\cite{malla2023drama} localizes risk and provides natural-language captions for safety-critical cues. To explicitly structure multi-stage reasoning, DriveLM~\cite{sima2024drivelm} proposes Graph-VQA with a VLM-based agent. DriveCoT~\cite{wang2024drivecot,dosovitskiy2017carla} curates chain-of-thought traces in CARLA to enhance interpretability of E2E policies. Reason2Drive~\cite{nie2024reason2drive} further targets interpretable, chain-based reasoning tasks tailored to AD.
Percept-WAM \emph{retains the general reasoning competence} of pretrained VLMs, while \emph{strengthening spatial perception} via World-PV/World-BEV tokens.

%% file: sec/3_methods.tex
\vspace{-2mm}
\section{Method}
\label{sec:method}
\vspace{-1mm}
\noindent\textbf{Overall Architecture of Percept-WAM}: As shown in \Cref{fig: architecture}, Percept-WAM contains: 1) the VLM backbone (\textit{i.e.}, InternVL2-8B~\cite{chen2024expanding}) to maintain general reasoning capability, 2) the learnable BEV-Level grid tokens to implicitly model the mapping from PV features to BEV-space representations, and
3) an additional action expert head for efficient and accurate trajectory decoding. Note that these BEV-level grid tokens can be initialized from point cloud features produced by a pretrained LiDAR encoder~\cite{lang2019pointpillars}, if LiDAR modality is available.

\noindent\textbf{Summary of the Tasks.}
As shown in Table \ref{tab:summary_tasks},
Percept-WAM accepts multi-view streaming video, LiDAR point (opt.), and textual queries as inputs. It supports \textbf{PV Perception} (2D detection, instance segmentation, semantic segmentation, mono-3D detection), \textbf{BEV Perception} (BEV 3D detection, BEV segmentation), and \textbf{Trajectory Prediction}.

\subsection{World-PV: Perspective-View Perception}
\label{sec:pv}


\begin{figure}[t!]
  \centering
  \begin{subfigure}[t]{\linewidth}
    \centering
    \includegraphics[width=\linewidth]{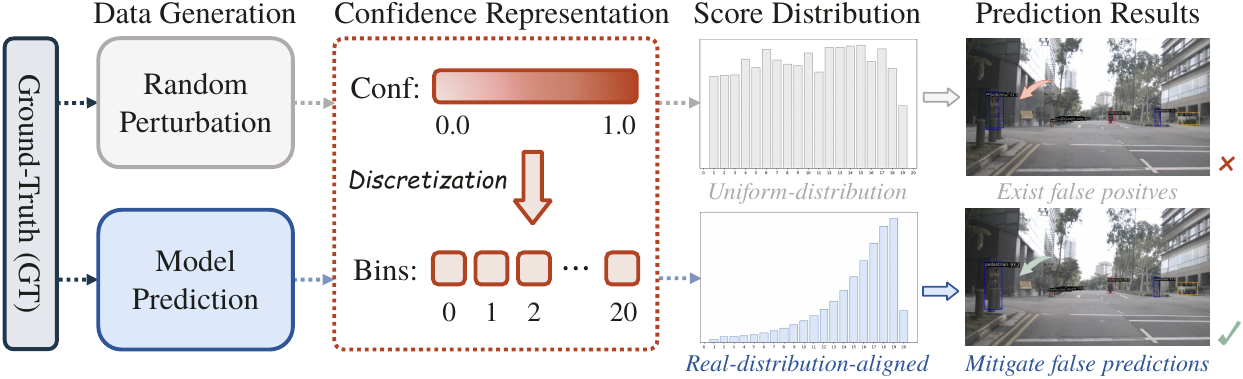}
    \caption{Different Ways of Confidence-tuning Dataset Generation.}
    \label{fig:pv_data_generator}
  \end{subfigure}
  \hfill
  \begin{subfigure}[t]{\linewidth}
    \centering
    \includegraphics[width=\linewidth]{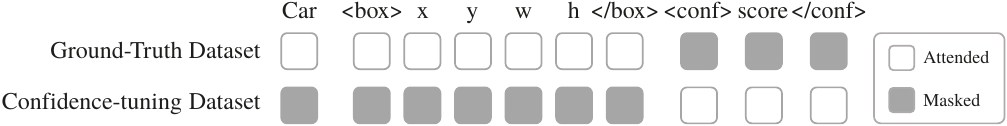}
    \caption{Loss-Mask of IoU-based Confidence Learning during Training.}
    \label{fig:pv_loss_mask}
  \end{subfigure}
  \vspace{-2mm}
  \caption{\textbf{Illustration of IoU-based confidence training strategy.}
  (a) The confidence-tuning dataset is generated by model predictions on the GT dataset. This yields scores that better match real distributions and reduce false positives compared to training on a random perturbation dataset. 
  (b) During training, different dataset strategies are supervised through a loss-mask scheme that promotes precise box and confidence tokens learning.}
  \label{fig:PV_detection}
  \vspace{-4mm}
\end{figure}

The perspective-view (PV) branch is a critical component of Percept-WAM for detecting objects across varying distances and scales. Distant vehicles and small or irregularly shaped objects are challenging to detect due to limited visual information. To overcome these difficulties, we support high-resolution input that preserves fine-grained details at large distances and efficient parallel AR decoding. This approach ensures high throughput while enabling reliable detection of objects under long-tail conditions.

As shown in \Cref{fig: architecture}, to leverage the pre-trained VLM (\textit{i.e.}, InternVL~\cite{chen2024expanding})’s understanding of image structure and semantics, the image inputs are first encoded by the VLM backbone of Percept-WAM.
Then the obtained image features, denoted as World-PV tokens, are  patchified into an $H{\times}W$ grid, and each grid location acts as a localized query for single-object perception.
Inspired by UFO~\cite{tang2025ufo},
we construct the grid tokens by interpolating from the World-PV tokens, yielding fine-grained features tied to local image coordinates.
Finally, each grid token predicts the bounding box or segmentation mask aligned with its coordinates, supervised by the corresponding ground truth.

\subsubsection{Task Formulation}
\label{sec:task_formulation}

\noindent\textbf{2D Detection and Monocular 3D  Detection.} We retain the conventional language-based autoregressive (AR) decoding paradigm, serializing the output predictions into natural-language–like token sequences~\cite{vaswani2017attention}.
Specifically for 2D detection in PV space, we formulate the output as
\vspace{-2mm}
\begin{center}
\textit{$cls$,\texttt{<}box\texttt{>}$x$,$y$,$w$,$h$\texttt{</}box\texttt{>},\texttt{<}conf\texttt{>}$s$\texttt{</}conf\texttt{>}}. 
\end{center}
\vspace{-2mm}
where $cls$, $s$ denote the object category and confidence score, $(x,y)$ and $(w,h)$ are the box center and size.
For 3D detection, the output follows the same sequence format, while the \textit{\texttt{<}box\texttt{>}} field is instead defined as $x,y,z,w,h,\ell,\theta,v_x,v_y$, representing the 3D center $(x,y,z)$, box size $(w,h,\ell)$, yaw angle $\theta$ and horizontal and vertical velocities $(v_x,v_y)$.
Following sequence-based detection methods~\cite{chen2021pix2seq}, continuous labels (e.g., coordinates and confidence) are normalized and discretized into integer bins, and supervised with cross-entropy loss~\cite{chen2021pix2seq}. 
Since categories are provided as text, the detector naturally supports open-vocabulary detection~\cite{li2022grounded,minderer2022simple}, which improves robustness in long-tail road scene understanding.
Besides, Percept-WAM decodes predictions in parallel across grids, significantly enhancing inference efficiency without sacrificing perception accuracy.

\noindent\textbf{2D Instance and Semantic Segmentation.}
Similar to UFO’s formulation~\cite{tang2025ufo}, we cast segmentation as feature retrieval task without adding new parameters: the model predicts $K{=}16$ \textit{\texttt{<}MASK\texttt{>}} tokens, and we retrieve masks by dot-product similarity between World-PV tokens and the $K$ mask tokens.
The generated mask is then interpolated to match the output resolution, with masks for all categories produced in a single forward pass.

\noindent\textbf{High-Resolution Input.}
Following InternVL-style dynamic tiling~\cite{chen2024far,chen2024expanding}, we split high-resolution images into non-overlapping tiles, each encoded using shared ViT weights. The features of these tiles are then fused through global positional alignment, allowing us to preserve long-range detail while avoiding quadratic memory growth.

\subsubsection{IoU-based Confidence Prediction}
\label{sec:pv-confidence}

Training–inference mismatch in MLLMs~\cite{jiang2025detect} often yields duplicate boxes in perception tasks.
Previous approaches, such as UFO~\cite{tang2025ufo} derive box confidence from the softmax of class logits.
However, large language models are systematically overconfident~\cite{zhang2024calibrating}: softmax probabilities saturate even for ambiguous detections, yielding many false positives especially in duplicate scenes.
Therefore, we add an \emph{IoU-based confidence token} for each predicted box, conditioned on its label attributes (class/coordinates/size).
This is aligned with prior quality-aware scoring~\cite{jiang2018acquisition,li2020generalized} and yields a more interpretable, localization-sensitive confidence.

\begin{itemize}
\item \textbf{Confidence-tuning Dataset Generation.}
To support training with IoU-based confidence supervision,
we build an auxiliary  set with IoU score from the training set (in \Cref{fig:pv_data_generator}).
We use the model from an intermediate training stage to inference on the training images, then the predicted boxes that match GT become samples paired with their IoU.
We discretize IoU to $20$ bins for WAM’s token-learning scheme.
Compared with random-offset synthesis from GT (which produces near-uniform IoU), the model-prediction distribution better reflects realistic confidence and reduces false detections. Refer to \Cref{sec:ablations} for more details.

\item \textbf{Confidence Learning during Training.}
We mix GT data and the confidence-tuning data during training.
As shown by the loss mask in \Cref{fig:pv_loss_mask},
for GT samples (IoU fixed to $1$), the model learns class and box without explicit IoU supervision to avoid collapse.
For confidence-tuning samples, the model predicts IoU conditioned on the perturbed labels, and only the loss for confidence is applied.

\item \textbf{Confidence Computation during Inference.}
During inference, the final detection score is defined as the product of class confidence (the softmax of the class token) and the predicted IoU score, providing a more unified and interpretable reliability measure.
\end{itemize}

To summarize the PV perception losses, 
we adopt token-level cross-entropy on discretized labels for object detection, following Pix2Seq~\cite{chen2021pix2seq}, and a combination of cross-entropy, sigmoid focal loss~\cite{lin2017focal}, and Dice loss~\cite{milletari2016v} for instance and semantic segmentation.

\subsection{World-BEV: BEV-View Perception}
\label{sec:bev}

\begin{figure}[t!]
  \centering
    \centering
    \includegraphics[width=0.8\linewidth]{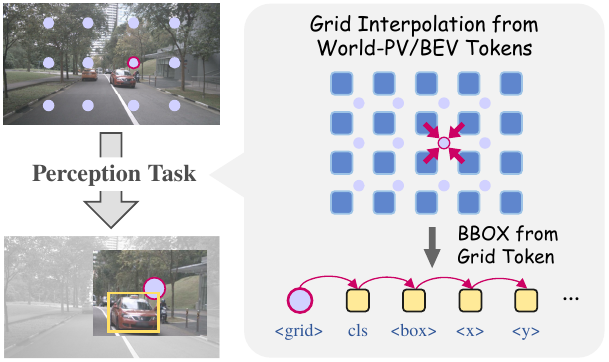}
    \vspace{-2mm}
  \caption{
  \textbf{Illustration of grid query tokens in dense prediction.}
  Note that the grid tokens are interpolated from World-PV or World-BEV tokens to predict the matched bounding box.
  }
  \label{fig:bev}
  \vspace{-4mm}
\end{figure}

3D spatial understanding is the cornerstone of reliable autonomous driving systems. 
To enhance the 3D perception capabilities of our Percept-WAM model, we explicitly integrate 3D object detection and semantic map segmentation tasks within the BEV representation space, enabling comprehensive scene understanding through multi-task learning.
Specifically, 
we propose World-BEV tokens, which are a set of learnable query tokens that serve as the foundation for BEV perception. 
Similar to previous work ~\cite{liu2022bevfusion,li2024bevformer,huang2021bevdet}, we instantiate the BEV space as an $H \times W$ grid centered on the ego vehicle.
Each token (\textit{i.e.}, grid cell) outputs a high-dimensional embedding, providing sufficient capacity to encode the spatial and semantic information, such as objects and map elements.
These World-BEV tokens are then used to query the features of the World-PV tokens established in Section~\ref{sec:pv} via cross-attention, enabling the model to lift 2D evidence into a 3D BEV representation in a purely data-driven manner.

The design of World-BEV tokens has two major advantages:
(i) Like World-PV tokens, World-BEV tokens can also be computed in a single forward pass in the prefilling stage, delivering high inference efficiency for trajectory prediction.
(ii) World-BEV tokens can be easily adapted for seamless multi-modal sensor fusion.
Under camera-only input, the word embeddings of these tokens are randomly initialized and optimized in the training stage.
When other 3D sensor inputs, such as LiDAR, are available, we use an additional encoder to extract point cloud features and use them to initialize the World-BEV tokens.
In particular, we first encode the point cloud with PointPillars~\cite{lang2019pointpillars} and then downsample the features via PixelUnshuffle~\cite{shi2016real} and an MLP layer.
This strategy injects pretrained, metrically grounded 3D priors into the BEV representation, boosting geometric consistency and accuracy.

\noindent\textbf{BEV 3D Object Detection.} 
For the 3D object detection task, 
we represent each 3D bounding box as a plain-text sequence:
\textit{cls, \texttt{<}box\texttt{>} x, y, z, w, h, l, $\theta$, $v_x$, $v_y$ \texttt{<}/box\texttt{>}}.
The continuous value of each attribute is normalized and discretized into integers within $[0, 1024)$.
As shown in \Cref{fig:bev}, 
we uniformly sample grid queries from World-BEV tokens via bilinear interpolation.
Then, by controlling the attention mask only to the relevant tokens (shown in Appendix \ref{sec:more_methods}), the object proposals are predicted independently in a parallel AR decoding manner from the grid query token and World-BEV tokens.

\noindent\textbf{BEV Map Segmentation.}
For the segmentation task, we reuse the PV segmentation formulation with 16 \textit{\texttt{<}mask\texttt{>}} tokens (see Section~\ref{sec:task_formulation} for more details). Similarly, we compute dot-product similarity scores between each World-BEV token and all \textit{\texttt{<}mask\texttt{>}} tokens, and then interpolate the predicted mask to match the desired output resolution. It should be noted that  different map categories may overlap (e.g., crosswalk is a subset of drivable area). Therefore, we cast map segmentation as independent binary segmentation tasks for each class.
In order to train the BEV tasks, we use cross-entropy loss (CE)~\cite{chen2021pix2seq} for BEV detection, and a combination of CE, sigmoid focal loss~\cite{lin2017focal}, and Dice loss~\cite{milletari2016v} for BEV map segmentation.

\subsection{From Perception to Action}
\label{sec:e2e}

\begin{figure}[t!]
    \centering
    \includegraphics[width=\linewidth]{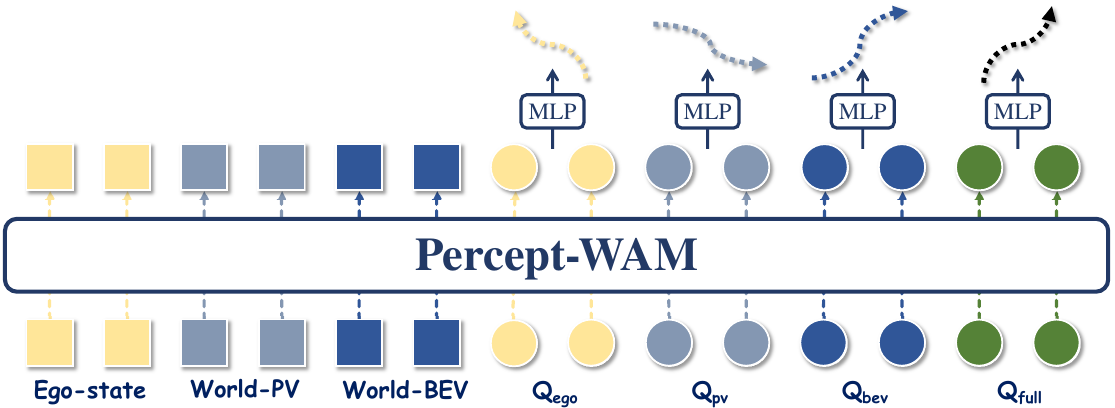}
  \caption{
  \textbf{Trajectory decoding.} Four sets of point-level queries interact with different input modality information, and generate trajectory using MLP. $\mathbf{Q}_\text{ego}$, $\mathbf{Q}_\text{pv}$, and $\mathbf{Q}_\text{bev}$ are aligned with their corresponding modality tokens via attention masking, while $\mathbf{Q}_\text{full}$ accesses all features to decode the final trajectory. }
  \label{fig:traj_method}
  \vspace{-4mm}
\end{figure}

\begin{table*}[t!]
  \centering
  \caption{Comparison of PV 2D and 3D perception performance on autonomous driving (AD) and general (Gen) datasets. Note that ‘Det' and ‘Seg’ denote Detection and Segmentation tasks respectively. The model trained on both AD and the general dataset, shown in the last row of the table, is the final \textbf{Percept-WAM} model. \textbf{Bold} emphasizes top method; \underline{underline} indicates the runner-up.}
  \vspace{-2mm}
  \label{tab:pv}
  \resizebox{\linewidth}{!}{
  \begin{tabular}{@{}lccccccccccc@{}}
\toprule
\multirow{3}{*}{\textbf{Method}} & \multirow{3}{*}{\begin{tabular}[c]{@{}c@{}}\textbf{Visual}\\ \textbf{Backbone}\end{tabular}} & \multirow{3}{*}{\textbf{LLM}} & \multicolumn{2}{c}{\textbf{2D Det}}       & \multicolumn{2}{c}{\textbf{Mono3D Det}}              & \multicolumn{2}{c}{\textbf{2D Instance Seg}}       & \multicolumn{3}{c}{\textbf{2D Semantic Seg}}                       \\ \cmidrule(l){4-12} 
                        &                                  &                      & nuImages      & COCO          & \multicolumn{2}{c}{nuScenes}           & nuImages      & COCO          & nuImages      & ADE20K        & COCOstuff    \\ \cmidrule(l){4-12} 
                        &                                  &                      & mAP           & mAP           & mAP                    & NDS           & mAP           & mAP           & mIoU          & mIoU          & mIoU          \\ \midrule
Mask R-CNN~\cite{he2017mask}              & RN101-FPN                        & --                    & --             & 39.8          & --                      & --             & --             & 37.1          & --             & --             & --             \\
Pix2Seq v2~\cite{chen2022unified}              & ViT-B                            & --                    & --             & 46.5          & --                      & --             & --             & 38.2          & --             & --             & --             \\
GiT~\cite{wang2024git}                     & ViT-B                            & --                    & --             & 46.7          & --                      & --             & --             & 31.9 & --             & 47.8          & --             \\
DINO~\cite{zhang2022dino}                    & RN50                             & --                    & --             & \underline{49.4} & --                      & --             & --             & --          & --             & --             & --             \\
Mask R-CNN~\cite{he2017mask}              & RN50                             & --                    & 47.8          & --             & --                      & --             & \underline{38.6}          & --             & --             & --             & --             \\
FCOS3D~\cite{wang2021fcos3d}                  & RN101 w/FT                 & --                    & --             & --             & 32.1                   & \textbf{39.5} & --             & --             & --             & --             & --             \\
Mask2Former~\cite{cheng2022masked}             & Swin-B                           & --                    & --             & --             & --                      & --             & --             & --             & --             & 52.4 & --             \\
DeepLab V2~\cite{caesar2018coco}              & VGG-16                           & --                    & --             & --             & --                      & --             & --             & --             & --             & 47.7          & \underline{33.2} \\ \midrule
Griffon-G-27B~\cite{zhan2024griffon}           & CLIP-ViT-L                       & Gemma2-27B           & --             & 40.6          & --                      & --             & --             & --             & --             & --             & --             \\
Groma~\cite{ma2024groma}                   & DINOv2                           & Vicuna-7B            & --             & 43.6          & --                      & --             & --             & --             & --             & --             & --             \\
VLM-FO1-3B~\cite{liu2025vlm}              & DaViT-L                          & QwenVL2.5-3B        & --             & 44.4          & --                      & --             & --             & --             & --             & --             & --             \\
VisionLLM~\cite{wang2023visionllm}               & RN50                             & Alpaca-7B            & --             & 44.8          & --                      & --             & --             & 25.2          & --             & --             & --             \\
LMM-Det~\cite{li2025lmm}                 & OWLv2-L                          & Vicuna-7B            & --             & 47.5       & --                         & --             & --             & --             & --             & --             & --             \\
UFO-internvl2-8B~\cite{tang2025ufo}        & InternViT                        & InternVL2-8B         & 13.7           & 48.9             & --                      & --             & 13.1           & \underline{42.6}             & 8.9          & \textbf{54.5}             & 30.2             \\
\textbf{Percept-WAM}{\small (2D AD)}      & InternViT                        & InternVL2-8B         & 46.7          & --    & --                      & --    & 36.8          & --             & \textbf{64.7} & --             & --             \\
\textbf{Percept-WAM}{\small (AD)}         & InternViT                        & InternVL2-8B         & \textbf{49.9} & --             & \underline{32.6} & 38.0          & \textbf{41.7} & --             & \underline{63.9} & --             & --             \\
\textbf{Percept-WAM}{\small (AD+Gen)} & InternViT                        & InternVL2-8B         & \underline{49.6} & \textbf{51.7} & \textbf{33.0}            & \underline{38.6}          & \textbf{41.7} & \textbf{45.9} & 62.8          & \underline{54.3} & \textbf{50.3} \\ \bottomrule
\vspace{-6mm}
\end{tabular}
}
\end{table*}

By introducing World-PV and World-BEV tokens, our model gains strong 2D perception and 3D scene understanding capability. Building on this, we introduce World-Action tokens to enable future trajectory generation for autonomous driving. We adopted a query-based trajectory decoding approach, following the paradigm of imitation learning for training. 
To enhance the alignment between the World-Action Tokens and tokens of different modalities, we introduce four sets of point-level queries via parallel decoding under controlled attention mask for efficient training.

\begin{itemize}
\item
    \textbf{Perception-Action Alignment.} 
    The BEV information (\textit{i.e.}, World-BEV tokens) captures accurate dynamic and static context, images (\textit{i.e.}, World-PV tokens) provide rich semantic details, and ego-state data supply vehicle kinematic information.
    Reliable future trajectory prediction requires features from all three perspectives. Therefore, 
    as shown in \Cref{fig:traj_method}, we set up four sets of point-level queries: \( \mathbf{Q}_{\mathbf{pv}} \), \( \mathbf{Q}_{\text{bev}} \),  \( \mathbf{Q}_{\text{ego}} \) and \( \mathbf{Q}_{\text{full}} \). The first three queries interact only with their corresponding modality features by controlling the attention mask, while \( \mathbf{Q}_{\text{full}} \) has access to all features. This ensures that the action is fully aligned with different input features, avoiding over-reliance on any single modality.
\item
    \textbf{Parallel Trajectory Decoding.} 
    Each query set \( \mathbf{Q} \in \mathbb{R}^{N \times C}\), where N represents the number of trajectory points and C is the dimensionality of the features, is randomly initialized. The query features are then encoded by Percept-WAM and decoded through an MLP to obtain the trajectory.
    During training, the four sets of queries are decoded in parallel to generate trajectories and supervised using Smooth-L1 loss~\cite{girshick2015fast}.
    During inference, the trajectory decoded from \( \mathbf{Q}_{\text{full}} \) is used as the final output.
\item 
    \textbf{Streaming Inference.} To further improve efficiency, we incorporate a streaming KV cache strategy into Percept-WAM. To mitigate distribution drift caused by training-inference paradigm mismatch, we adopt a longer-clip training scheme and a dual-recomputation KV cache mechanism. Details are provided in Appendix~\ref{sec:stream}.
\end{itemize}

\begin{table}[t!]
  \centering
  \caption{Summary of tasks and corresponding training datasets used in Percept-WAM training. Unless otherwise specified, we follow the official train-val-test dataset split.}
  \small
  \vspace{-2mm}
  \setlength{\tabcolsep}{2pt}
  \resizebox{\linewidth}{!}{
  \begin{tabular}{lll}
    \toprule
    \textbf{Perception} &  \textbf{Tasks}  & \textbf{Datasets}\\
    \midrule
    \multirow{5}{*}{\textbf{PV}}
      & 2D Det & nuImages~\cite{caesar2020nuscenes}, nuScenes~\cite{caesar2020nuscenes}, COCO~\cite{lin2014microsoft} \\
      & Mono 3D Det & nuScenes, Waymo~\cite{sun2020scalability} \\
      & Instance  Seg & nuImages, COCO \\
      & Semantic Seg & \makecell[l]{nuImages, ADE20K~\cite{zhou2019semantic}, COCOStuff~\cite{caesar2018coco}}\\
      & Grounding &  \makecell[l]{RefCOCO~\cite{kazemzadeh2014referitgame}, RefCOCO+~\cite{yu2016modeling}, \\RefCOCOg~\cite{mao2016generation}}\\
    \midrule
    \multirow{2}{*}{\textbf{BEV}} 
      & 3D Det &  \multirow{2}{*}{\makecell[l]{nuScenes, Waymo, nuScenes Map~\cite{caesar2020nuscenes}}} \\
      & BEV Seg & \\
      \midrule
    \textbf{Trajectory}
      & Waypoint Pred & nuScenes, NAVSIM~\cite{dauner2024navsim} \\
    \midrule
    \textbf{Driving QA} & QA & \makecell[l]{DriveLM-nuscenes~\cite{sima2024drivelm}, LingoQA~\cite{marcu2023lingoqa}, \\CODA-LM~\cite{li2024automated}, Dolphins~\cite{ma2024dolphins}, IDKB~\cite{lu2025can}, \\MapLM~\cite{cao2024maplm}, DriveGPT4~\cite{xu2024drivegpt4}} \\
    \bottomrule
  \end{tabular}
  }
  \vspace{-4mm}
  \label{tab:summary_tasks}
\end{table}

%% file: sec/4_experiments.tex
\vspace{-2mm}
\section{Experiments}
\vspace{-1mm}
\label{sec:experiments}

\subsection{Experimental Setup}

\begin{table*}[ht]
\centering
\caption{Comparison of trajectory planning methods on nuScenes and NAVSIM~\cite{dauner2024navsim} benchmarks.
$\downarrow$ indicates lower is better, $\uparrow$ indicates higher is better.
Our Percept-WAM matches or surpasses existing trajectory planning method. Two-stage training (Percept-WAM*) further shows that a stronger perception-enhanced model achieves better downstream E2E trajectory planning performance.}
\vspace{-2mm}
\resizebox{0.9\textwidth}{!}{%
\begin{tabular}{lcccccccccc}
\toprule
\multirow{2}{*}{\textbf{Method}} &
\multicolumn{4}{c}{\textbf{nuScenes}} &
\multicolumn{6}{c}{\textbf{NAVSIM v1}} \\
\cmidrule(lr){2-5} \cmidrule(lr){6-11}
 & L2-1s$\downarrow$ & L2-2s$\downarrow$ & L2-3s$\downarrow$ & Avg.$\downarrow$ 
 & NC$\uparrow$ & DAC$\uparrow$ & TTC$\uparrow$ & Comf.$\uparrow$ & EP$\uparrow$ & PDMS$\uparrow$ \\
\midrule
UniAD~\cite{hu2023planning} & 0.20 & 0.42 & 0.75 & 0.46 & 97.8 & 91.9 & 92.9 & 100 & 78.8 & 83.4 \\
VAD-Base~\cite{jiang2023vad} / VAD-v2~\cite{chen2024vadv2} & 0.17 & 0.34 & 0.60 & 0.37 & 97.2 & 89.1 & 91.6 & 100 & 76.0 & 80.9 \\
DiffusionDrive~\cite{liao2025diffusiondrive} & 0.27 & 0.54 & 0.90 & 0.57 & 98.2 & 96.2 & 94.7 & 100 & 82.2 & 88.1 \\
DriveVLM~\cite{tian2024drivevlm} & 0.18 & 0.34 & 0.68 & 0.4 & -- & -- & -- & -- & -- & -- \\
BEV-Planner~\cite{li2024ego} & \textbf{0.16} & \textbf{0.32} & \textbf{0.57} & \textbf{0.35} & -- & -- & -- & -- & -- & -- \\
DRAMA~\cite{yuan2024drama} & -- & -- & -- & -- & 98.0 & 93.1 & \textbf{94.8} & 100 & 80.1 & 85.5 \\
Hydra-MDP~\cite{li2024hydra} & -- & -- & -- & -- & 98.3 & 96.0 & \underline{94.6} & \textbf{100} & 78.7 & 86.5 \\
\midrule
\textbf{Percept-WAM} & \underline{0.17} & 0.35 & 0.63 & 0.38 & \underline{98.7} & \underline{97.8} & 93.2 & 92.8 & \underline{84.4} & \underline{88.6} \\
\textbf{Percept-WAM*} & \textbf{0.16} & \underline{0.33} & \underline{0.60} & \underline{0.36}{\scriptsize\,(+5.2\%)} & \textbf{98.8} & \textbf{98.6} & 94.4 & 99.5 & \textbf{84.8} & \textbf{90.2}{\scriptsize\,(+1.6)} \\
\bottomrule
\end{tabular}
\label{tab:nuscenes_navisim}
}
\vspace{-4mm}
\end{table*}

We train the Percept-WAM from InternVL2-8B~\cite{chen2024expanding} with various training datasets as shown in \Cref{tab:summary_tasks}. 
The training process applies AdamW optimizer \cite{loshchilov2017decoupled} with the base LR $0.0002$ and weight decay $0.01$. We train the model in cosine decay learning rate schedule with linear warmup for $1000$ steps. Mixed precision \cite{micikevicius2017mixed} and gradient checkpointing \cite{chen2016training} are enabled to save GPU memory. 
World-PV is discretized into a $10\times10$ grid for both detection and segmentation, while World-BEV uses the $40\times40$ grid for detection and a $10\times10$ grid for segmentation to provide finer spatial granularity in BEV space.
We adopt a two-stage curriculum that first consolidates spatial grounding for PV/BEV perception and then aligns the planner through E2E VLA fine-tuning. 
Further hyper-parameters and detailed data composition can be found in Appendix \ref{sec:more_experiment_settings}.

\subsection{Main Results}

\begin{figure}[t!]
  \centering
  \includegraphics[width=\linewidth]{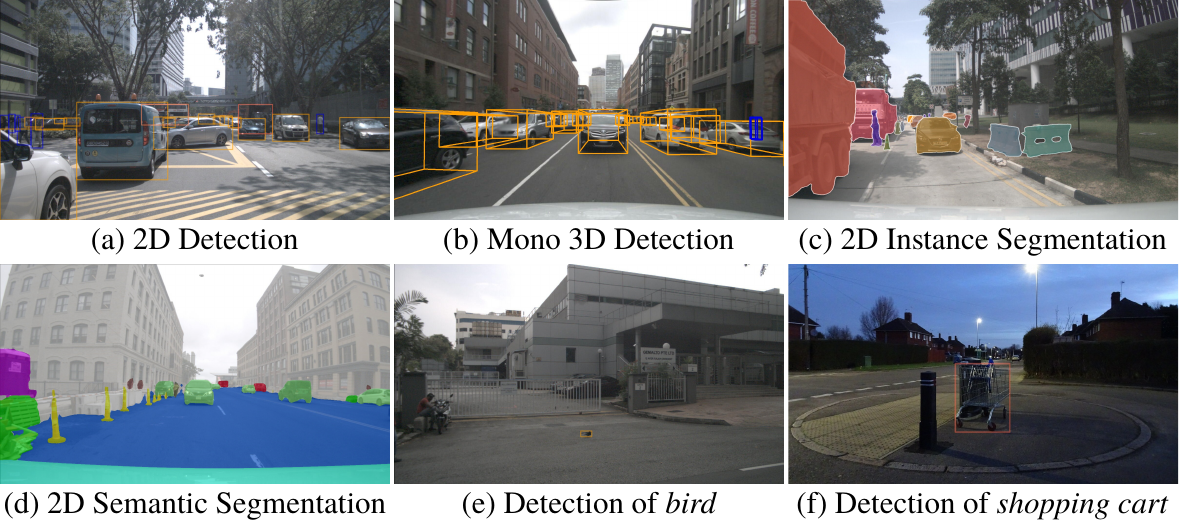}
  \vspace{-6mm}
  \caption{
  \textbf{PV perception visualization}. 
  Percept-WAM demonstrates i) accurate and stable performance in crowded, long-range, and small-object scenarios for autonomous driving (see (a)–(d)); and ii) robust open-vocabulary perception in general-domain tasks (see (e) and (f)). Note that detection boxes of the same category are labeled with the same color.
  }
  \label{fig:pv-qual}
  \vspace{-2mm}
\end{figure}

\noindent\textbf{PV Perception Results.} As shown in \Cref{tab:pv}, Percept-WAM matches or surpasses specialist detectors and segmentors on both the \emph{nuImages}~\cite{caesar2020nuscenes} and \emph{nuScenes}~\cite{caesar2020nuscenes} PV tasks. For instance, it attains $49.9$ mAP in 2D detection and $41.7$ mAP in 2D instance segmentation compared to $47.8$ and $38.6$ mAP with Mask R-CNN~\cite{he2017mask}, and $33.0$ mAP in mono 3D detection versus $32.1$ mAP with FCOS3D~\cite{wang2021fcos3d}.
We observe a synergy between 2D and 3D PV detection, with 2D detection performance improving by $3.2$ mAP. This gain results from unified 2D and 3D modeling, as shown by joint training on AD datasets, where training across all PV tasks leads to consistent gains across benchmarks.
Additionally, training on AD and general-purpose datasets achieves performance comparable to or better than specialist and multimodal large language models in both autonomous driving and general scenarios.
Visualizations in \Cref{fig:pv-qual} demonstrate the excellent capability of Percept-WAM in detecting and segmenting multiple objects in complex scenarios.

\begin{table}[t!]
  \centering
  \caption{Results of BEV perception tasks on nuScenes \textit{\textbf{val}} dataset, where
  \textit{Dri.}→drivable area, \textit{Ped.}→pedestrian crossing, \textit{Lane.}→lane divider, \textit{Veh.}→vehicle.
  }
  \label{tab:bev-det-seg}
  \vspace{-2mm}
  \scalebox{0.85}{
  \begin{tabular}{lcccccc}
    \toprule
    \multirow{2}{*}{\textbf{Method}} & \multicolumn{2}{c}{\textbf{Detection}} & \multicolumn{4}{c}{\textbf{Segmentation (IoU)}} \\
    \cmidrule(l){2-3}
    \cmidrule(l){4-7}
     & \textbf{mAP} & \textbf{NDS} & \textbf{Dri.} & \textbf{Ped.} & \textbf{Lane.} & \textbf{Veh.} \\
    \midrule
    {\small BEVDet~\cite{huang2021bevdet}} & {\small 0.360} & {\small 0.438} & {\small --} & {\small --} & {\small --} & {\small --} \\
     {\small BEVFormer-S~\cite{li2024bevformer}} & {\small 0.375} & {\small 0.448} & {\small 80.7} & {\small --} & {\small 21.3} & {\small 43.2} \\
    {\small PointPillars~\cite{lang2019pointpillars}} & {\small 0.523} & {\small 0.613} &{\small --} &{\small --} &{\small --} &{\small --}\\
    {\small SECOND~\cite{yan2018second}} & {\small 0.526} & {\small 0.630} &{\small --} &{\small --} &{\small --} &{\small --}\\
    {\small BEVFusion~\cite{liu2022bevfusion}} & {\small {\textbf{0.685}}} & {\small {\textbf{0.714}}} &{\small {\underline{85.5}}} &{\small {\underline{60.5}}} & {\small {\textbf{67.7}}} &{\small --} \\
 \textbf{Percept-WAM} & {\small \underline{0.589}} &{\small \underline{0.645}} &{\small \textbf{87.0}} &{\small \textbf{70.9}} &{\small \underline{62.7}} &{\small \textbf{60.2}} \\
    \bottomrule
  \vspace{-6mm}
 \end{tabular}
 }
\end{table}

\begin{figure}[t!]
  \centering
  \includegraphics[width=\linewidth]{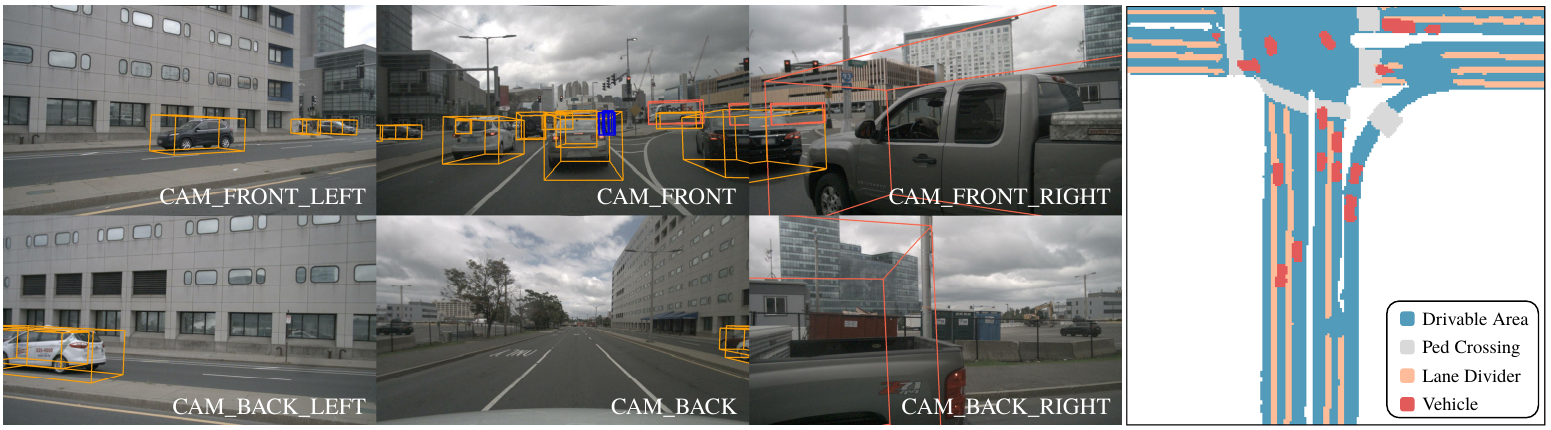}
  \vspace{-6mm}
  \caption{
  \textbf{BEV perception visualization}. Left: BEV 3D object detection results projected onto surrounding images; Right: BEV map segmentation for the same scene sample.
  }  
  \label{fig:bev-qual}
  \vspace{-2mm}
\end{figure}

\noindent\textbf{BEV Perception Results.}
As shown in \Cref{tab:bev-det-seg}, without using sequential information, and with a relatively low image input resolution (\textit{i.e.}, $448 \times 796$), Percept-WAM can outperform many specialist models on both detection and segmentation tasks.
Specifically, for map segmentation, Percept-WAM can achieve superior performance with BEVFusion \cite{liu2022bevfusion} on drivable area and pedestrian crossing.
Segmentation results on geographical train/val splits are summarized in Appendix \ref{sec: more results}.
For object detection, our method achieves an mAP of 0.589, which can outperform classical detectors such as PointPillars \cite{lang2019pointpillars} and SECOND \cite{yan2018second}.
Qualitative results on nuScenes val set are shown in \Cref{fig:bev-qual}.
Please note that, rather than outperforming state-of-the-art methods on each sub-task, the main purpose of integrating BEV perception tasks is to strengthen the 3D spatial understanding capabilities of Percept-WAM.

\begin{figure}[t!]
  \centering
  \includegraphics[width=\linewidth]{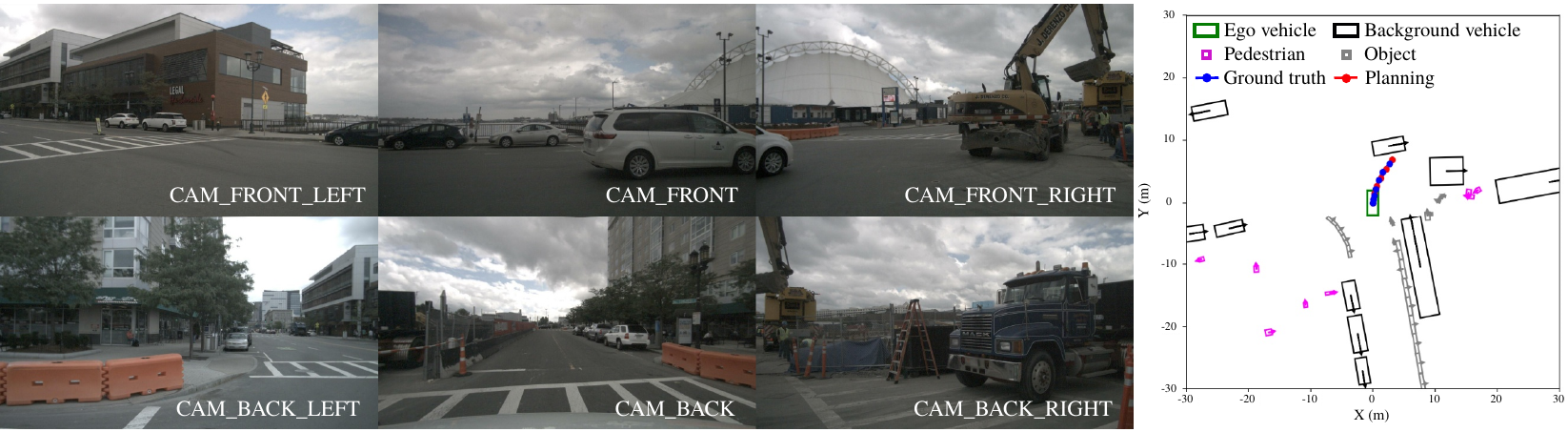}
  \vspace{-6mm}
  \caption{
      \textbf{Trajectory planning visualization.} Left: Input surrounding images; Right: Ego vehicle trajectory planning in BEV perspective. Our model successfully navigates a construction zone, yielding to an oncoming vehicle while making the right turn.
    }  
  \vspace{-4mm}
  \label{fig:traj_vis}
\end{figure}

\noindent\textbf{E2E Trajectory Planning Results.}
\label{sec:e2e_results}
\Cref{tab:nuscenes_navisim} shows E2E driving performance of different methods. Specifically, our model achieves an average trajectory L2 error of 0.36m on nuScenes' open-loop metrics and a score of 90.2 on NAVSIM's closed-loop metrics, outperforming most existing BEV-based methods like UniAD~\cite{hu2023planning} and VLM-based methods like DriveVLM~\cite{tian2024drivevlm}. Experimental results show that our two-stage training strategy further improves model performance.
For NAVSIM, we draw inspiration from Hydra-MDP~\cite{li2024hydra}, improving performance by scoring and selecting optimal trajectories from a static vocabulary. The visual results in \Cref{fig:traj_vis} further demonstrate the strong trajectory planning capability of the model, even in complex environments.

\subsection{Ablations on Different Settings}
\label{sec:ablations}

\noindent\textbf{Ablation on IoU-based Confidence Prediction.} 
To validate the impact of different IoU score distributions in confidence-tuning dataset (refer to Section~\ref{sec:pv-confidence}), 
we compare three construction schemes: (i) \textit{random-perturb}: perturb ground-truth (GT) boxes and uniformly sample boxes across IoU levels; (ii) \textit{uniform model-pred}: uniformly sample model predictions across IoU; and (iii) \textit{real model-pred}: directly use model predictions with their realistic IoU distribution.
As shown in \Cref{tab:pv-ablation}, the proposed IoU-based confidence prediction, aligned with the realistic distribution of model predictions, significantly improves performance by $1.5$ AP and $2.3$ AP$_{75}$, while ‘random-perturb’ and ‘uniform model-pred’ settings are inferior to the baseline.

\begin{table}[t!]
  \centering
  \caption{Ablation studies on training with different confidence-tuning datasets for IoU-based confidence prediction, evaluated on the 2D detection task of the nuImages \textit{\textbf{val}} set.}
  \label{tab:pv-ablation}
  \vspace{-2mm}
  \resizebox{0.45\textwidth}{!}{%
  \begin{tabular}{lccc}
    \toprule
    \textbf{Confidence Computation} & \textbf{AP} & \textbf{AP$_{50}$} & \textbf{AP$_{75}$} \\
    \midrule
    \small Baseline (class score) & {\small 48.1} & {\small \textbf{70.9}} & {\small 51.4} \\
    \ \small + IoU Conf. (random-perturb) & {\small 46.9} & {\small 70.0} & {\small 50.7} \\
    \ \small + IoU Conf. (uniform model-pred) & {\small 46.2} & {\small 69.1} & {\small 49.3} \\
    \ \small + IoU Conf. (real model-pred) & {\small \textbf{49.6}} & {\small 70.4} & {\small \textbf{53.7}} \\
    \bottomrule
  \end{tabular}
  \vspace{-6mm}
  }
\end{table}

\begin{figure}[tb!]
  \centering
  \begin{subfigure}[t]{0.342\linewidth}
    \centering
    \includegraphics[width=\linewidth]{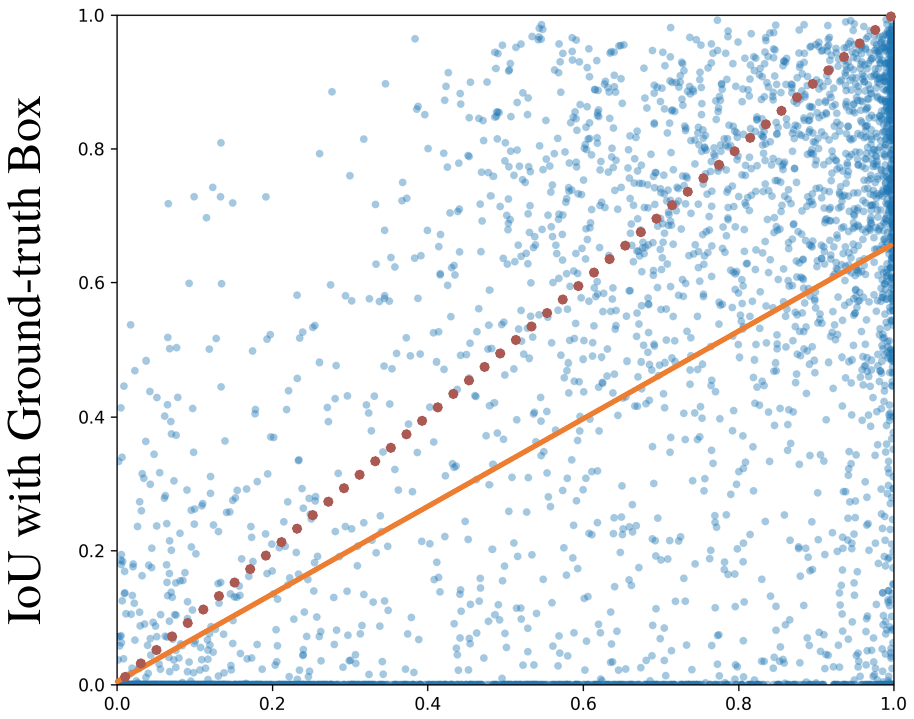}
    \caption{\centering \scriptsize Class score\\
                (baseline)}
    \label{fig:pv_abla_baseline}
  \end{subfigure}
  \begin{subfigure}[t]{0.312\linewidth}
    \centering
    \includegraphics[width=\linewidth]{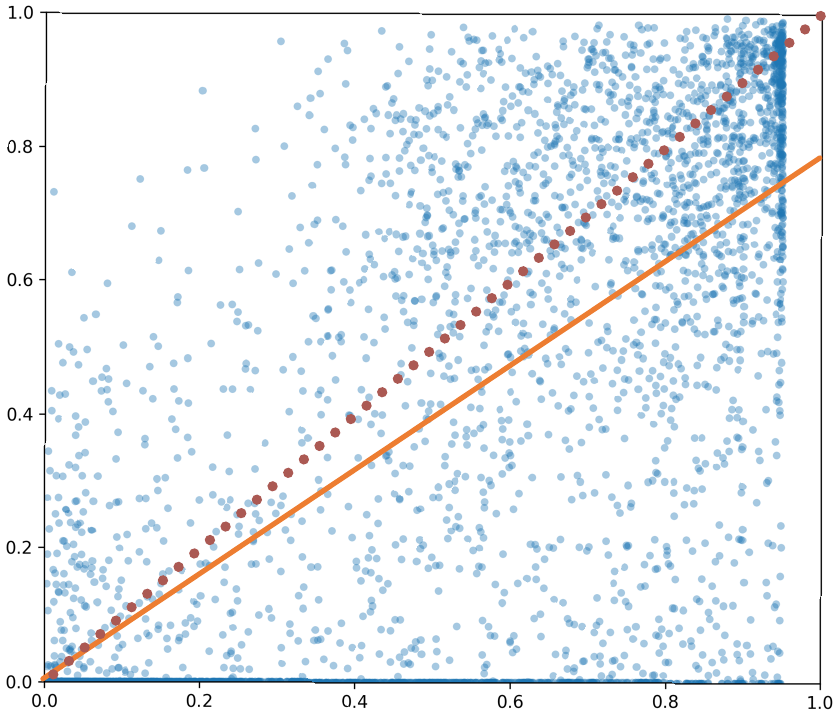}
    \caption{\centering \scriptsize Class score$\times$IoU conf\\
                (random-perturbation)}
    \label{fig:pv_abla_disturbBaseline}
  \end{subfigure}
  \begin{subfigure}[t]{0.312\linewidth}
    \centering
    \includegraphics[width=\linewidth]{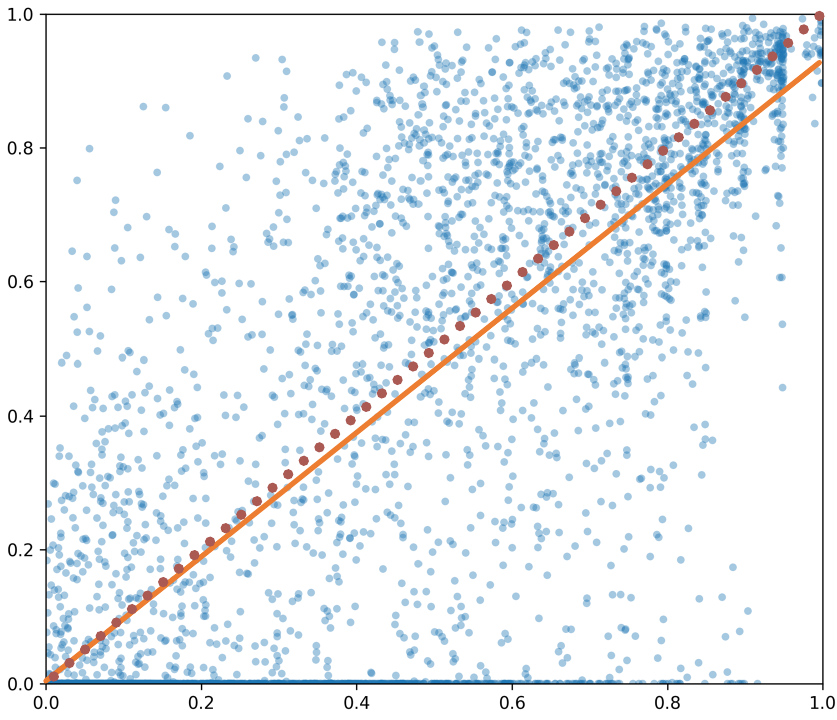}
    \caption{\centering \scriptsize Class score$\times$IoU conf\\
                (model-prediction)}
    \label{fig:pv_abla_disturbFromModel}
  \end{subfigure}
  \vspace{-2mm}
  \caption{
  \textbf{Ablation on IoU-based confidence prediction.}
  Each point denotes a detected bounding box. The dashed line ($y=x$) presents samples with precise score prediction.
  By multiplying the class score by IoU confidence, the fitting curve in (c) aligns closer to the diagonal than (a)/(b), indicating more accurate box scores.
  }
  \label{fig:pv_confidence}
  \vspace{-2mm}
\end{figure}

To further demonstrate the impact of confidence-score training on model confidence, we visualize the relationship between predicted confidence (x-axis) and the IoU of the corresponding boxes with ground-truth (y-axis) in \Cref{fig:pv_confidence}.
As shown in \Cref{fig:pv_abla_baseline}, for class-score only setting, a large number of low-quality bounding boxes with high confidence scores (lower-right region) remain unfiltered in post-processing, which degrades precision. 
After incorporating IoU-based confidence prediction, the points in \Cref{fig:pv_abla_disturbBaseline} and \Cref{fig:pv_abla_disturbFromModel} move closer to the line $y = x$.
The curve in \Cref{fig:pv_abla_disturbFromModel}, trained on the model prediction dataset, aligns more closely with the diagonal, indicating improved predicted scores accuracy better suited for post-processing.

\noindent\textbf{Ablation on BEV 3D Detection.} 
We conduct ablation studies on the nuScenes validation set to assess the impact of each component related with BEV detection task, with all results listed in \Cref{tab:bev-abl}.
Starting from a camera-only baseline, 
we initialize the World-BEV tokens with the feature extracted by a pretrained LiDAR encoder, which boosts the mAP by 8.2\%.
By adding data augmentations, such as LiDAR points and image data augmentation, the mAP is further improved by 8.1\%.
With an increase of the grid resolution (from $20 \times 20$ to $40 \times 40$), we observe an mAP improvement of 9.1\%.
Finally, we replace the AR coordinate decoding with MLP for parallel inference.
This maintains a 50.4 mAP but accelerates inference speed by 16$\times$.

\begin{table}[t!]
  \centering
  \caption{Ablations of BEV-perception related design choices on nuScenes \textit{\textbf{val}} set.}
  \vspace{-2mm}
  \label{tab:bev-abl}
  \resizebox{0.37\textwidth}{!}{%
  \begin{tabular}{lcc}
    \toprule
     \textbf{Method}& \textbf{ mAP}& \textbf{NDS} \\
    \midrule
     Baseline (Camera)& {\small 25.0}& {\small 25.7} \\
     \ \ + Lidar Encoder& {\small 33.2}& {\small 32.2} \\
     \ \ + Data Augmentation& {\small 41.3}& {\small 39.2} \\
    \ \ + Number of Sampling Grids & {\small 50.4}& {\small 46.6} \\
     \ \ + MLP Parallel (16× speedup)& {\small 50.4}& {\small 43.7} \\
 \bottomrule
  \end{tabular}
  \vspace{-4mm}
  }
\end{table}

\begin{table}[t!]
  \centering
  \caption{Ablations of different decoding mechanisms and efficiency improvement from streaming inference  on nuScenes \textit{\textbf{val}}.}
  \vspace{-2mm}
  \label{tab:traj-ablation}
  \resizebox{0.4\textwidth}{!}{%
  \begin{tabular}{lccc}
    \toprule
    \textbf{Decoding Mechanisms} & \textbf{L2 (avg.)$\downarrow$}  & \textbf{Latency (ms)$\downarrow$} \\
    \midrule
    AR  & {\small 0.3970} & {\small 2700}  \\
    \ \ + Streaming Infer.  & {\small 0.4058} & {\small 2209}  \\
    AR(cluster) & {\small 0.3919} & {\small 1470} \\
    Query-base & {\small 0.3822} & {\small 1174} \\
    \ \ + Streaming Infer.  & {\small 0.3839} & {\small 707}  \\
    \bottomrule
  \end{tabular}
  }
  \vspace{-4mm}
\end{table}

\noindent\textbf{Ablation on E2E training.}
Table \ref{tab:traj-ablation} presents the trajectory planning results of different decoding methods on the nuScenes validation set. ‘AR' refers to directly converting trajectories into text and predicting them autoregressively. ‘AR (cluster)' follows AutoVLA~\cite{zhou2025autovla}:  trajectories are first segmented and clustered to generate new action tokens, which are then autoregressively predicted and decoded into trajectories. The results show that our query-based approach achieves the best trade-off between accuracy and inference speed. Our streaming inference strategy provides further efficiency gains, reducing the inference time by 18\% and 40\% for  AR and query-based decoding approaches, respectively, with almost no impact on accuracy.

%% file: sec/5_conclusion.tex
\vspace{-2mm}
\section{Conclusion}
\vspace{-1mm}
This paper presents Percept-WAM, the first framework to unify 2D and 3D perception into one VLA model via \emph{World-PV} and \emph{World-BEV} tokens.
It can achieve performances on par with specialized models across all sub-tasks.
More importantly, by explicitly integrating these perception tasks into one unified model, the performance of trajectory planning can be further improved with high precision and low latency, demonstrating the effectiveness of our method.
In the future, we will explore offline/online RL with rollout-based rewards that couple perception accuracy with trajectory prediction, aiming to enforce the overall consistency.

%% file: sec/X_suppl.tex
\clearpage
\maketitlesupplementary

\section{More Method Details}

\subsection{The Attention Mask for Perception Tasks}
\label{sec:more_methods}
\begin{figure}[h!]
  \centering
  \begin{subfigure}[t]{0.44\linewidth}
    \centering
    \includegraphics[width=\linewidth]{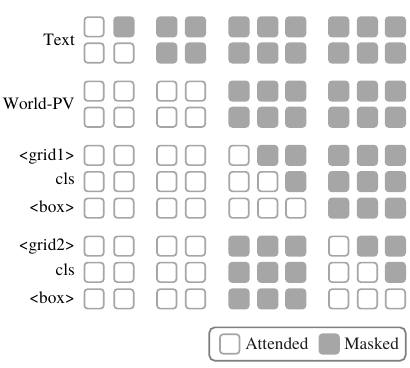}
    \caption{PV detection.}
    \label{fig:PV_detection_atten}
  \end{subfigure}
  \hfill
  \begin{subfigure}[t]{0.54\linewidth}
    \centering
    \includegraphics[width=\linewidth]{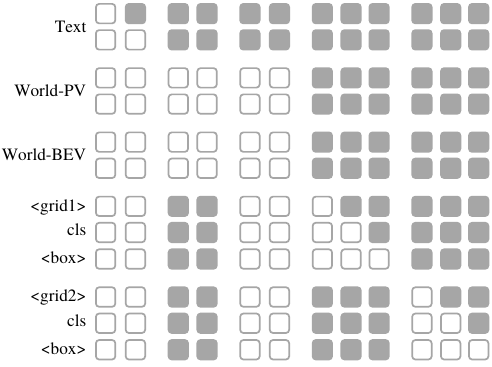}
    \caption{BEV detection.}
    \label{fig:BEV_detection_atten}
  \end{subfigure}
  \vspace{-2mm}
  \caption{Illustration of attention mask for input tokens for (a) PV and (b) BEV detection tasks.}
  \label{fig: grid attn mask}
  \vspace{-4mm}
\end{figure}

To illustrate how Percep-WAM leverages World-PV and World-BEV tokens in the unified VLM backbone, \Cref{fig: grid attn mask} visualizes the attention masks between input and output tokens for the PV- and BEV-detection tasks, respectively.

Specifically for the PV detection task, the attention mask is designed according to three principles:
(i) World-PV tokens are fully visible to each other to better fuse PV features;
(ii) for each grid-based prediction, the text tokens, grid tokens and output tokens follow the standard causal attention used in LLMs;
(iii) to support grid-based parallel AR decoding, grid tokens and output tokens from different grid-based predictions are mutually masked. For BEV detection, two design choices apply: (i) similar to World-PV tokens, World-BEV tokens are fully mutually visible; (ii) to better capture the PV-to-BEV transformation and mitigate overfitting, each output token is constrained to attend only to World-BEV tokens and its corresponding grid token.

\subsection{Streaming Inference}
\label{sec:stream}

\begin{figure}[t!]
  \centering
  \begin{subfigure}[t]{0.42\linewidth}
    \centering
    \includegraphics[width=\linewidth]{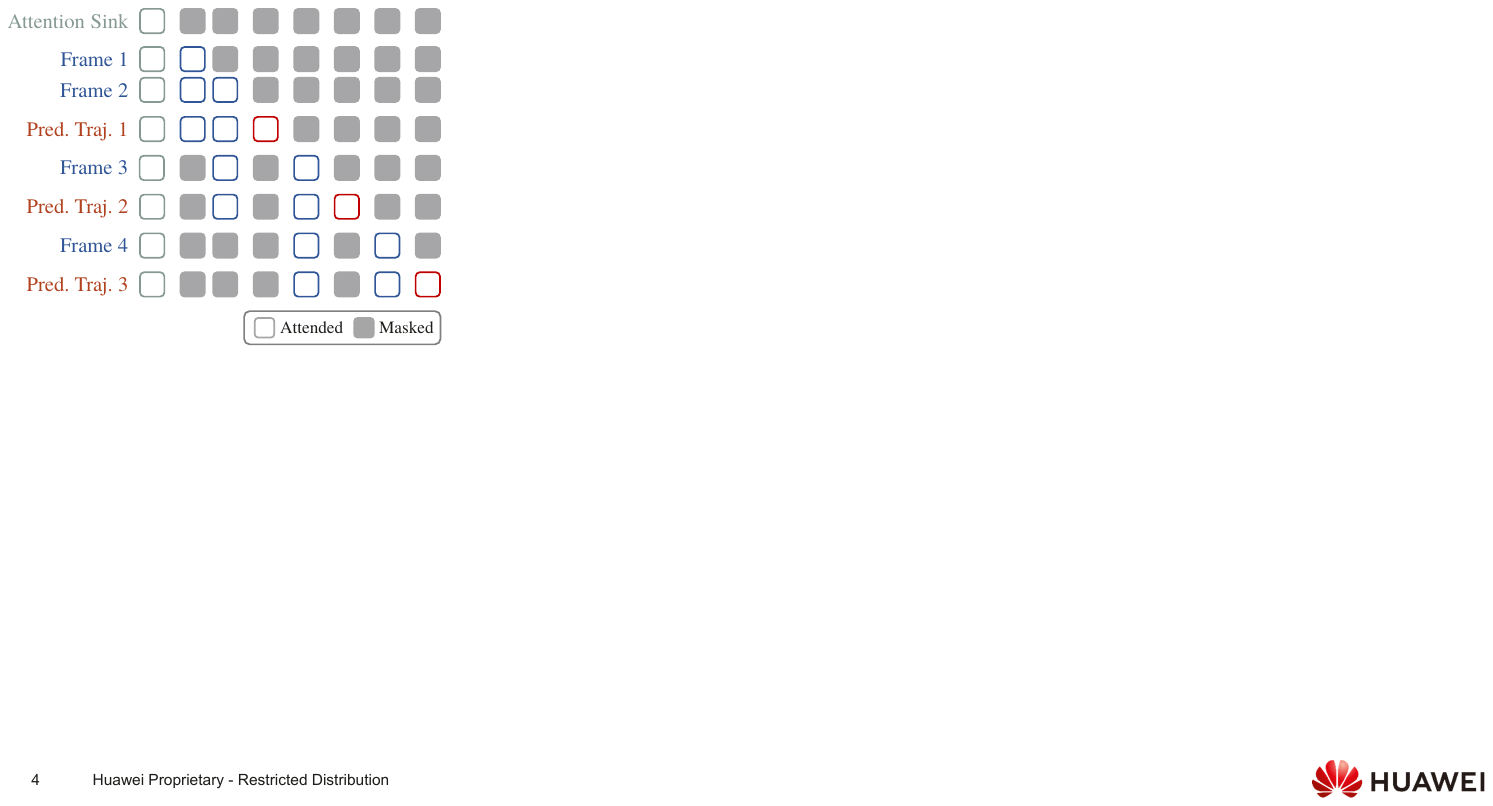}
    \caption{Streaming Attention Map.}
    \label{fig:stream attn map}
  \end{subfigure}
  \hfill
  \begin{subfigure}[t]{0.56\linewidth}
    \centering
    \includegraphics[width=\linewidth]{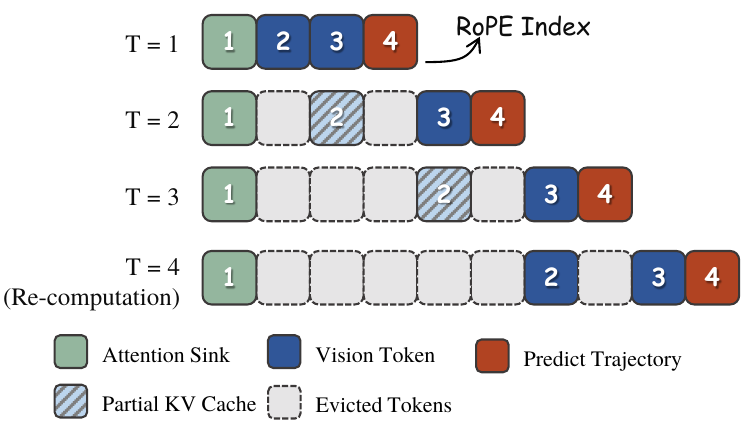}
    \caption{Streaming Strategy.}
    \label{fig:stream strategy}
  \end{subfigure}
  \vspace{-2mm}
  \caption{\textbf{Streaming inference strategy.} Each trajectory prediction attend previous two frames. The KV cache of historical frames (blue diagonally striped blocks in Fig.(b)) is reused to improve inference efficiency. 
  A dual-recomputation strategy is proposed to mitigate the effects of discontinuous cache positions and error accumulation.}
  \label{fig: inference attn mask}
  \vspace{-4mm}
\end{figure}

To meet the demands of real-world applications, we investigate the streaming inference capability of the Percept-WAM model for handling infinitely long conversational visual inputs.
Inspired by streaming inference research in current mainstream VLMs~\cite{xu2025streamingvlm, ning2025livevlm}, we adopt a streaming strategy illustrated in \Cref{fig: inference attn mask}. 
Specifically, as shown in \Cref{fig:stream attn map}, the trajectory prediction tokens at time step 
$T$ attend to the tokens of the two most recent frames (\textit{i.e.}, frame $T$ and frame $T\!-\!1$). Considering the high computational cost of processing visual tokens in the prefill phase, we reuse the KV cache from previous computations, with the specific strategy available for reference in \Cref{fig:stream strategy}.

However, inconsistencies between training and inference paradigms inevitably lead to \textit{distribution drift}: the cross-attention mechanism between frames theoretically enables historical tokens to implicitly encode historical information of infinite length. This differs significantly from standard video-clip-based training, which relies on fixed-length historical information. Thus, adopting a longer-clip training scheme is crucial for enhancing the model’s generalization capability on extended historical sequences.

To stabilize the computation process while gradually discarding historical visual tokens, we retain the attention sink~\cite{xiao2023efficient} (corresponding to the green grids in \Cref{fig:stream strategy}). However, this design causes deviations in positional embedding within the KV cache, arising from the discontinuity between the attention sink and visual frames. To address this issue, a dual-recomputation strategy is proposed with both local attention refinement and global cache recomputation: we recompute rotary positional embeddings (corresponding to the white numbers in \Cref{fig:stream strategy}) and adopt the specific token recomputation method~\cite{yao2025cacheblend} to correct cross-attention results. Meanwhile, we mitigate error accumulation from increasing inference lengths by periodically caching ViT tokens and recomputing the complete KV cache, safeguarding long-sequence stability.

As shown in \Cref{tab:traj-ablation}, we achieve 16\% and 40\% reduction in inference latency for the AR and the Query decoders, respectively, with an accuracy loss of less than 0.01.

\begin{figure}[t!]
  \centering
  \includegraphics[width=0.8\linewidth]{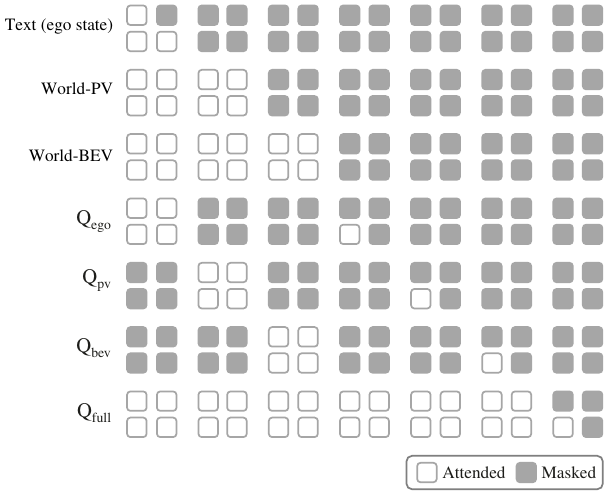}
  \caption{\textbf{Attention mask for our query-based trajectory decoding.} $\mathbf{Q}_\text{ego}$, $\mathbf{Q}_\text{pv}$, and $\mathbf{Q}_\text{bev}$ are aligned with their corresponding modality tokens, while $\mathbf{Q}_\text{full}$ accesses all features to decode the final trajectory.}  
  \vspace{-4mm}
  \label{fig:traj_attention}
\end{figure}

\subsection{End-to-End Trajectory Prediction}
\noindent\textbf{Attention Mask for Query-based Trajectory Decoding.} 
As described in Section~\ref{sec:e2e}, we introduce several sets of point-level queries and enforce modality-specific alignment by adjusting the attention mask. The attention mask is visualized in \Cref{fig:traj_attention}: the first three query sets attend only to their corresponding modality tokens for trajectory decoding, while the final query set attends to all input tokens to generate the final trajectory.

\noindent\textbf{Detail Settings of Percept-WAM as the Trajectory Selector.} 
As mentioned in Section~\ref{sec:e2e_results}, training Percept-WAM solely to replicate ground-truth trajectories is insufficient for achieving strong closed-loop performance, as trajectory supervision often misaligns with real-world evaluation metrics~\cite{witte2025epipolar}. To mitigate this gap, we introduce a query-based trajectory scoring and selecting approach inspired by Hydra-MDP~\cite{li2024hydra} and GTRS~\cite{li2025generalized}, evaluating it on the NAVSIM v1 benchmark. 

\begin{figure}[t!]
    \centering
    \includegraphics[width=0.85\linewidth]{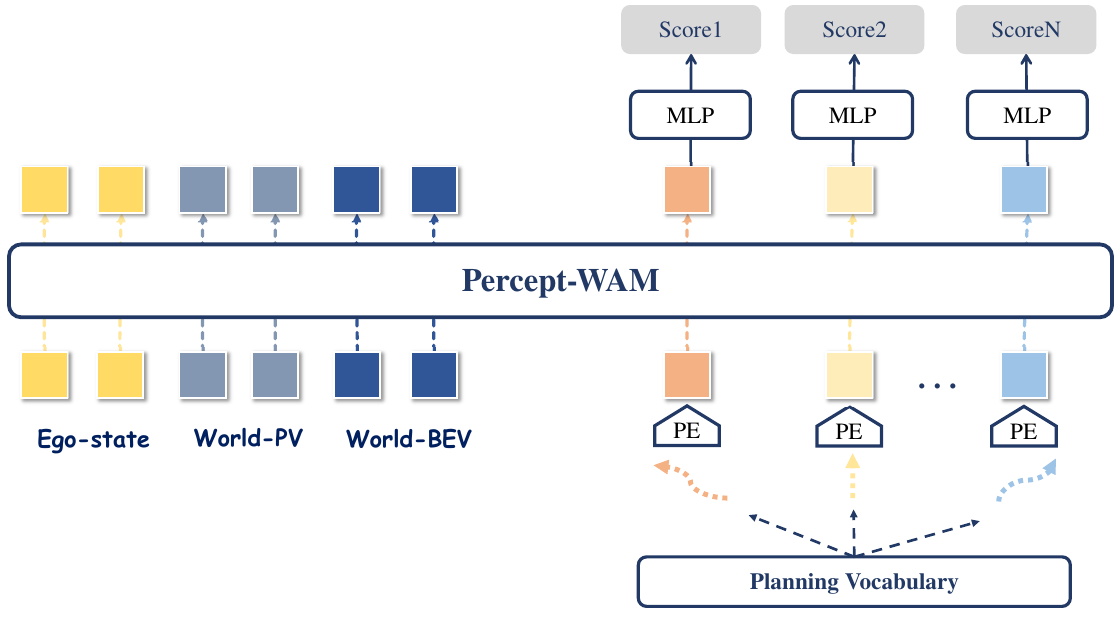}
  \caption{
  \textbf{Query-based trajectory scoring and selecting.} The embedded trajectory vocabularies are trained to be aligned with the World-PV, World-BEV, and Text Tokens, then decoded to generate the composite scores. The trajectory with the maximal score is selected as the planning result.}
  \label{fig:traj_method_sup}
  \vspace{-4mm}
\end{figure}

As showed in \Cref{fig:traj_method_sup}, instead of direct trajectory replication, we train the model to score a super-dense, pre-clustered trajectory vocabulary $\mathcal{V}_{XL}$. Half of $\mathcal{V}_{XL}$ is randomly dropped during training to improve robustness. At inference, a reduced vocabulary $\mathcal{V}_{L}$ is used. Each trajectory in $\mathcal{V}_{L}$ is first embedded through an MLP and then encoded as $\mathcal{V}^\prime_L$ via a stack of transformer layers:

\vspace{-2mm}
{\small
\begin{equation}
\mathcal{V}^\prime_L=\textit{Transformer}(Q,K,V=\textit{MLP}(\mathcal{V}_{L})).
\end{equation}
}

Percept-WAM further integrates contextual cues by querying features from the World-PV Tokens $T_{\text{WPV}}$, World-BEV Tokens $T_{\text{WBEV}}$ and Text Tokens $T_{\text{T}}$, generating World-Action Tokens $T_{\text{WA}}$ for trajectory scoring: 

{\small
\vspace{-2mm}
\begin{equation}
T_{\text{WA}}=\textit{Percept-WAM}(Q=\mathcal{V}^\prime_L,K,V=T_{\text{WPV}}, T_{\text{WBEV}}, T_{\text{T}}).
\end{equation}
\vspace{-4mm}
}

These enriched features are passed through a set of prediction heads to compute trajectory scores. Using a binary cross-entropy objective, we distill rule-based driving priors into our model. At inference time, our model selects the trajectory with the highest composite score, reflecting optimal performance for the given scenario.

\begin{table*}[t!]
  \centering
  \caption{Hyperparameter configurations for Percept-WAM across training stages. Note that within each stage, jointly trained tasks share the same base hyperparameters. All experiments are trained on 8 nodes.}
  \setlength{\tabcolsep}{6pt}
  \setlength{\extrarowheight}{3pt}
  \resizebox{0.9\textwidth}{!}{%
  \begin{tabular}{lcccc}
    \toprule
    & \multicolumn{3}{c}{\textbf{Stage 1}} & \textbf{Stage 2} \\
    \cmidrule(lr){2-4}
    \textbf{Training parameter} 
      & \textbf{PV Perception} 
      & \textbf{BEV Perception}  
      & \textbf{Driving QA}  
      & \textbf{Trajectory Prediction} \\
    \midrule
    Learning rate 
      & 0.0002 & 0.0002 & 0.0002 & 0.0002 \\
    Warmup iterations 
      & 1000 & 1000 & 1000 & 500 \\
    Training iterations 
      & 100000 & 100000 & 100000 & 3000 \\
    Batch size 
      &  64 & 64 & 64 & 64 \\ 
    Image resolution 
      & $1344\times896$ & $796\times448$ & $796\times448$ & $796\times448$ \\
    Grid number 
      & $10\times10$ & $40\times40$ for Det, $10\times10$ for Seg & NA & NA \\  
    Optimizer 
      & AdamW & AdamW & AdamW & AdamW \\  
    Weight decay 
      & 0.01 & 0.01 &  0.01 & 0.01 \\  
    Schedule 
      & Cosine Annealing & Cosine Annealing & Cosine Annealing & Cosine Annealing\\  
    \bottomrule
  \end{tabular}
  }
  \label{tab:hyperparameter}
\end{table*}

\begin{table*}[t!]
  \centering
  \caption{Comparison of referring expression comprehension (REC) performance, reported using P@0.5. VGM and MLLM refer to the vision generalist model and multimodal large language model, respectively.}
  \label{tab:rec}
  \resizebox{0.73\textwidth}{!}{%
\begin{tabular}{@{}lcccccccccc@{}}
\toprule
\multirow{2}{*}{\textbf{Method}} & \multirow{2}{*}{\textbf{Type}} & \multicolumn{3}{c}{\textbf{RefCOCO}}                & \multicolumn{3}{c}{\textbf{RefCOCO$+$}}                     & \multicolumn{2}{c}{\textbf{RefCOCOg}}    &             \\ \cmidrule(l){3-10} 
                        &                       & \textbf{val}            & \textbf{testA}          & \textbf{testB}          & \textbf{val}            & \textbf{testA}          & \textbf{testB}          & \textbf{val}            & \textbf{test}           & \multirow{-2}{*}{\textbf{Avg}}            \\ \midrule
MDETR~\cite{kamath2021mdetr}                   & \multirow{2}{*}{VGM}  & 86.8          & 89.6          & 81.4          & 79.5          & 84.1          & 70.6          & 81.6          & 80.9          & 81.8          \\
Grounding DINO T~\cite{liu2024grounding}       &                       & 89.2 & \textbf{91.9} & 86.0          & 81.1          & 87.4          & 74.7          & 84.2          & 84.9          & 84.9          \\ \midrule
Shikra-13B~\cite{chen2023shikra}              & \multirow{4}{*}{MLLM} & 87.8          & 91.1          & 81.8          & 82.9          & 87.8          & 74.4          & 82.6          & 83.2          & 84.0          \\
MiniGPT-v2-7B~\cite{chen2023minigpt}           &                       & 88.1          & 91.3          & 84.3          & 79.6          & 85.5          & 73.3          & 84.2          & 84.3          & 83.8          \\
VistaLLM-7B~\cite{pramanick2024jack}             &                       & 88.1          & 91.5          & 83.0          & 82.9          & \textbf{89.8} & 74.8          & 83.6          & 84.4          & 84.8          \\
\textbf{Percept-WAM}     &                       & \textbf{89.9}          & 90.8          & \textbf{89.3} & \textbf{85.4} & 88.2          & \textbf{81.7} & \textbf{86.5} & \textbf{87.0} & \textbf{87.4} \\ \bottomrule
\end{tabular}}
\end{table*}

\begin{table*}[t!]
  \centering
  \caption{Comparison of referring expression segmentation (RES) performance, reported using cumulative IoU (cIoU). VGM and MLLM refer to the vision generalist model and multimodal large language model, respectively.}
  \label{tab:res}
  \resizebox{0.73\textwidth}{!}{%
\begin{tabular}{@{}lcccccccccc@{}}
\toprule
\multirow{2}{*}{\textbf{Method}} & \multirow{2}{*}{\textbf{Type}} & \multicolumn{3}{c}{\textbf{RefCOCO}}                & \multicolumn{3}{c}{\textbf{RefCOCO$+$}}                     & \multicolumn{2}{c}{\textbf{RefCOCOg}}    &             \\ \cmidrule(l){3-10} 
                        &                       & \textbf{val}            & \textbf{testA}          & \textbf{testB}          & \textbf{val}            & \textbf{testA}          & \textbf{testB}          & \textbf{val}            & \textbf{test}           & \multirow{-2}{*}{\textbf{Avg}}            \\ \midrule
GLEE-Pro~\cite{wu2024general}                 & & 80.0 & -- & -- & 69.6 & -- & -- & 72.9 & -- & 74.2 \\
UNINEXT-H~\cite{yan2023universal} & \multirow{-2}{*}{VGM}  & 82.2 & 83.4 & 81.3 & 72.5 & 76.4 & 66.2 & 74.7 & 76.4 & 76.6 \\ \midrule
LISA-7B~\cite{lai2024lisa} & & 74.1 & 76.5 & 71.1 & 62.4 & 67.4 & 56.5 & 66.4 & 68.5 & 67.9 \\
VistaLLM-13B~\cite{pramanick2024jack} &  & 77.2 & 78.7 & 73.9 & 71.8 & 74.4 & 65.6 & 69.8 & 71.9 & 72.9 \\
GLaMM-7B~\cite{rasheed2024glamm} & & 79.5 & 83.2 & 76.9 & 72.6 & 78.7 & 64.6 & 74.2 & 74.9 & 75.6 \\
HiMTok-8B~\cite{wang2025himtok} & & 81.1 & 81.2 & 79.2 & 77.1 & 78.8 & 71.5 & 75.8 & 76.7 & 77.7 \\
\textbf{Percept-WAM}      & \multirow{-5}{*}{MLLM} & \textbf{86.5} & \textbf{87.4} & \textbf{86.6} & \textbf{79.9} & \textbf{83.6} & \textbf{75.2} & \textbf{81.3} & \textbf{81.9} & \textbf{82.8} \\ \bottomrule
\end{tabular}}
\vspace{-2mm}
\end{table*}

\section{More Experiment Settings}
\label{sec:more_experiment_settings}

\subsection{Two-Stage Training Details.
}
\label{sec: two-stage training parameters}

\noindent\textbf{Training Setting Details.} To effectively optimize Percept-WAM for both perception and planning, we adopt a two-stage training scheme. The first stage focuses on enhancing the VLM’s overall 2D and 3D spatial perception, assisted by autonomous-driving general QA tasks, while the second stage trains end-to-end trajectory prediction on top of this perception-enhanced base model. The detailed training hyperparameters are summarized in \Cref{tab:hyperparameter}.

\noindent\textbf{Data Construction.}
As shown in \Cref{tab:summary_tasks} of the main paper, we employ task-specific training datasets for PV, BEV, and trajectory prediction. To preserve the model’s general capabilities, we further incorporate autonomous-driving QA data during training. For each task family, we specify the data mixture as follows: (i) for PV tasks, the sampling ratio among 2D Detection, Mono 3D Detection, Instance Segmentation, Semantic Segmentation, and Grounding is set to 1:1:2:1:1; (ii) for BEV tasks, the ratio between 3D Detection and BEV Segmentation is 1:1; (iii) when jointly training multiple main tasks (\textit{i.e.}, PV, BEV, trajectory prediction, and Driving QA), we sample each task with equal probability for simplicity.

\begin{table*}[t!]
\centering
\caption{Ablation studies of different trajectory decoding and selecting methods on NAVSIM~\cite{dauner2024navsim} v1 benchmarks.
$\downarrow$ indicates lower is better, $\uparrow$ indicates higher is better.
Query-based trajectory scoring and selecting surpasses other trajectory planning mechanisms on the NAVSIM v1 benchmark.}
\vspace{-2mm}
\begin{tabular}{lcccccc}
\toprule
\multirow{2}{*}{\textbf{Trajectory Planning Mechanism}} &
\multicolumn{6}{c}{\textbf{NAVSIM v1}} \\
\cmidrule(l){2-7} 
 & NC$\uparrow$ & DAC$\uparrow$ & TTC$\uparrow$ & Comf.$\uparrow$ & EP$\uparrow$ & PDMS$\uparrow$ \\
\midrule
AR-based trajectory generation & 96.4 & 94.5 & 92.0 & 98.7 & 78.5 & 84.1 \\
Query-based trajectory generation & 96.5 & 90.3 & 91.0 & \textbf{99.7} & 75.6 & 80.4 \\
Query-based trajectory scoring and selection & \textbf{98.8} & \textbf{98.6} & \textbf{94.4} & 99.5 & \textbf{84.8} & \textbf{90.2} \\
\bottomrule
\end{tabular}
\label{tab:ablation_navisim}
\vspace{-2mm}
\end{table*}

\section{More Experiment Results}

\subsection{Main Results}
\label{sec: more results}
 
\begin{figure*}[ht!]
    \centering
    \includegraphics[width=0.8\linewidth]{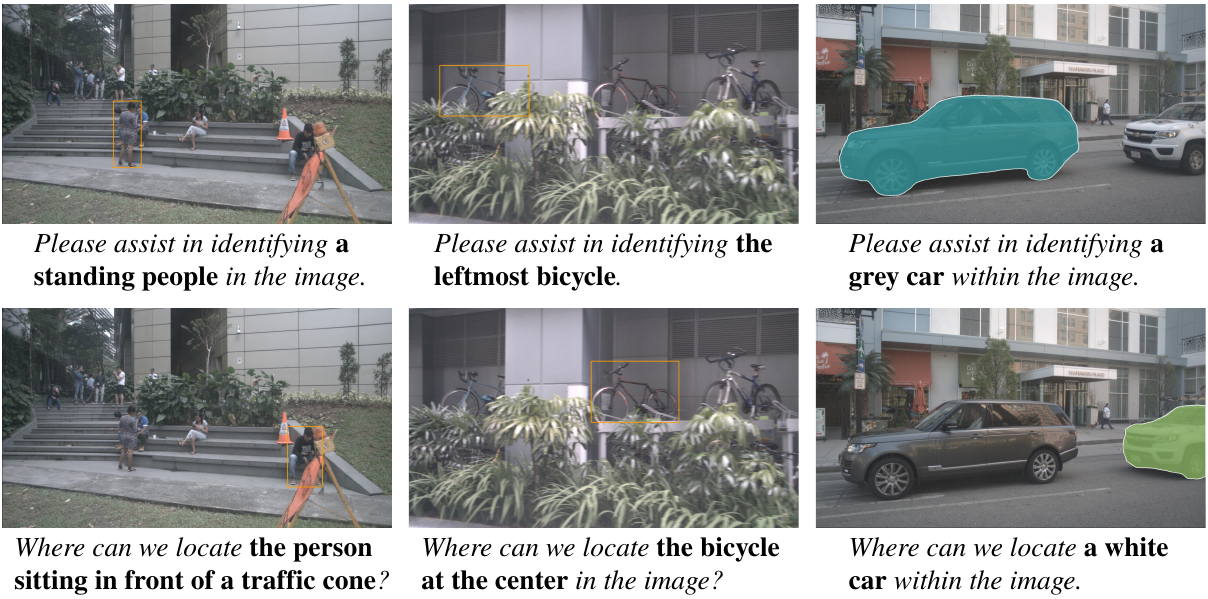}
    \vspace{-2mm}
  \caption{
  \textbf{Illustration of Percept-WAM on the visual grounding task.}
Percept-WAM accurately localizes referred objects and exhibits robust understanding of diverse visual attributes.
  }
  \label{fig:pv_rec}
\end{figure*}

\noindent\textbf{Comparison of Visual Grounding Performance.} 
In this section, we evaluate the model's ability to leverage fine-grained world features for complex visual grounding. Visual grounding is a critical task that associates textual descriptions with corresponding image regions or objects. This task can be further divided into referring expression comprehension (REC) and referring expression segmentation (RES), with output formats of bounding boxes or masks. We present comprehensive comparison results for both tasks in \Cref{tab:rec} and \Cref{tab:res}, respectively. As shown in \Cref{tab:rec}, Percept-WAM achieves top-tier performance on the RefCOCO~\cite{kazemzadeh2014referitgame}, RefCOCO$+$~\cite{yu2016modeling}, and RefCOCOg~\cite{mao2016generation} benchmarks among MLLMs, outperforming Grounding DINO~\cite{zhang2022dino}, a representative VGM, by an average of $2.5$ P@0.5. \Cref{tab:res} demonstrates Percept-WAM's exceptional pixel-level segmentation performance among VGMs and MLLMs, achieving an average cIoU of $82.8$. We demonstrate Percept-WAM's visual grounding performance on the same images with different descriptions, as shown in \Cref{fig:pv_rec}, highlighting its robust ability to distinguish specific objects with unique attributes from visually similar instances.

\noindent\textbf{BEV Segmentation Results on Geographical Train/Val Splits.} For the BEV segmentation task, we additionally report results on nuScenes with the geographical train/val split (following the split protocol in EAFT~\cite{witte2025epipolar}), as shown in \Cref{tab:bev-seg}. The results indicate that our Percept-WAM achieves performance comparable to EAFT, and that incorporating LiDAR inputs consistently provides stable performance gains.

\begin{table}[t!]
  \centering
  \caption{BEV segmentation results on nuScenes~\cite{caesar2020nuscenes} with geographical train/val splits~\cite{witte2025epipolar}.}
  \label{tab:bev-seg}
  \resizebox{0.48\textwidth}{!}{%
  \begin{tabular}{llcccc}
    \toprule
    \textbf{Modality}&\textbf{Method} & \textbf{Dri.}& \textbf{Ped.}& \textbf{Lane.}&\textbf{Veh.}\\
    \midrule
 \multirow{2}{*}{\textbf{C}}& {\small EAFT~\cite{witte2025epipolar}}& {\small 58.06}& {\small --}  & {\small --}  
&{\small --}  
\\
 & \textbf{Percept-WAM} & {\small 56.41}& {\small 33.21}& {\small 22.03}&{\small 34.47}\\
 \midrule
 \textbf{C + L}& \textbf{Percept-WAM} & {\small 67.19}& {\small 22.06}& {\small 23.99}&{\small 56.76}\\
 \bottomrule
    \end{tabular}}
    \vspace{-2mm}
\end{table}

\subsection{Ablation Studies}
\noindent\textbf{Ablation of Query-based Trajectory Decoding Methods.}
We ablate different query–mask configurations for trajectory decoding, as summarized in  \Cref{tab:query_mask_abs__results}. Compared to the “full” mode, which uses a single query set $\mathbf{Q}_{\mathbf{full}}$, parallel decoding with multiple query sets reduces the trajectory error by approximately 8\%.

\noindent\textbf{Ablation of Clustered-Action Design.}
The clustered-action method discretizes trajectories using the K-Disk clustering algorithm with a maximum vocabulary size of $2048$. A greedy clustering approach is applied based on a 0.05-meter distance threshold to group similar trajectories. To improve long-horizon stability, trajectories are segmented into \( s \)-frame intervals, where \( s \in \{1, 2, 3, 6\} \). As shown in \Cref{tab:ar_cluster_results}, a segment length of $2$ frames with a 0.05-meter threshold yields the best performance on nuScenes, achieving an L2\_avg of $0.3919$.

\begin{table}[t!]
  \centering
  \caption{Query mask ablation results on nuScenes trajectory prediction task. Lower is better. ‘Full' means using a single query set which can attend to all the input tokens. ‘Parallel' refers to our approach, where all query sets are decoded in parallel. }
  \vspace{-1mm}
  \label{tab:query_mask_abs__results}
  \begin{tabular}{cc}
    \toprule
    \textbf{Mask mode} &  \textbf{L2-avg$\downarrow$} \\
    \midrule
    Full & 0.4151 \\
    Parallel & \textbf{0.3821} \\
    \bottomrule
  \end{tabular}
  \vspace{-2mm}
\end{table}

\begin{table}[t!]
  \centering
  \caption{Ablation of AR (cluster) configurations with different frame interval $s$ 
on nuScenes trajectory prediction task. Lower($\downarrow$) is better.}
  \vspace{-1mm}
  \label{tab:ar_cluster_results}
  \resizebox{0.45\textwidth}{!}{%
  \begin{tabular}{lcccc}
    \toprule
    \textbf{Method} & \textbf{L2-1s$\downarrow$} & \textbf{L2-2s$\downarrow$} & \textbf{L2-3s$\downarrow$} & \textbf{L2-avg$\downarrow$} \\
    \midrule
    AR & 0.160 & 0.356 & 0.674 & 0.3970 \\
    AR(cluster)$_{s=1}$ & 0.434 & 0.744 & 1.129 & 0.7692 \\
    AR(cluster)$_{s=2}$ & 0.181 & 0.356 & 0.638 & \textbf{0.3919} \\
    AR(cluster)$_{s=3}$ & 0.196 & 0.390 & 0.692 & 0.4260 \\
    AR(cluster)$_{s=6}$ & 0.236 & 0.494 & 0.892 & 0.5406 \\
    \bottomrule
  \end{tabular}
  }
  \vspace{-2mm}
\end{table}

\noindent\textbf{Comparison of E2E Performance.} 
We ablate different trajectory planning strategies in the closed-loop setting on NAVSIM benchmark. 
Specifically, we compare three methods: (i) AR-based trajectory generation, (ii) query-based trajectory generation, (iii) query-based trajectory scoring and selection. 
The results are summarized in \Cref{tab:ablation_navisim}.

The comparison between AR-based and query-based methods in \Cref{tab:traj-ablation} appears inconsistent with the ablation results in \Cref{tab:ablation_navisim}. This discrepancy arises from the misalignment between open-loop and closed-loop metrics. While the query-based method reduces the L2 distance between planned trajectories and ground truth (from $1.1$ m to $0.8$ m on the NAVSIM navtrain validation split), improved trajectory replication does not always lead to better closed-loop performance, as evidenced by PDMS on the NAVSIM~\cite{dauner2024navsim} benchmark. Therefore, we adopt query-based trajectory scoring and selection which combines the imitation strength of the query-based approach with the closed-loop metrics, and achieves the best overall performance.  

\subsection{Illustrations}
\label{sec: more illustrations}

\begin{figure*}[ht!]
  \vspace{-2mm}
  \centering
  \includegraphics[width=0.8\linewidth]{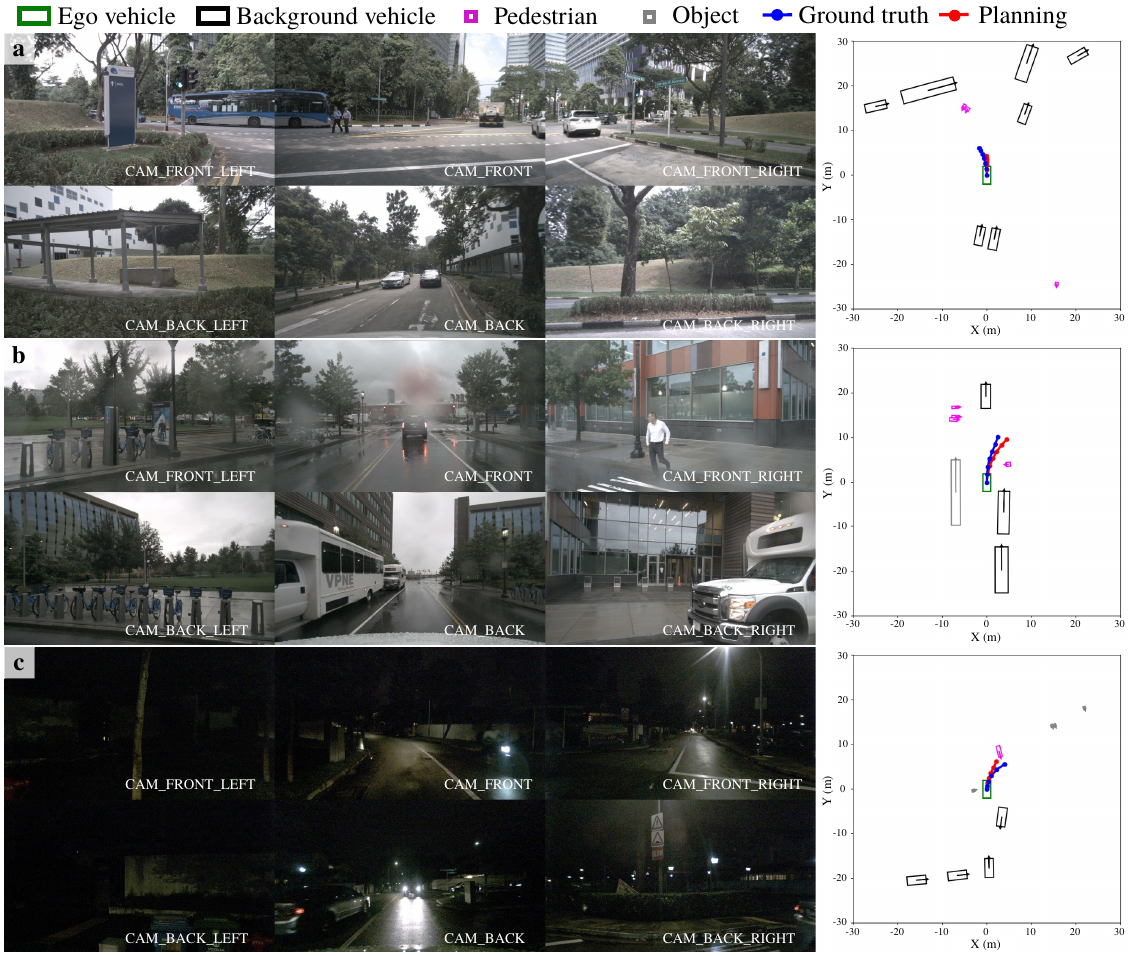}
    \vspace{-2mm}
  \caption{\textbf{More illustration of trajectory planning capabilities.} The ego vehicle is shown as a green box, background vehicles as black boxes, pedestrians as purple boxes, and other detected objects as grey boxes. Ground truth trajectories are shown as blue dots, and planned trajectories as red dots. (a) Percept-WAM successfully navigates the ego vehicle turning left, yielding to pedestrians crossing the road. (b) Under rainy conditions, the model detects a jaywalking pedestrian and safely avoids a potential collision. (c) While turning right at night, the model accounts for a cyclist obstructed in the front camera view, ensuring enough space for safe passage. Overall, our model demonstrates strong planning performance, even in challenging and long-tail scenarios.} 
  \label{fig: e2e_vis_append}
  \vspace{-3mm}
\end{figure*}

\noindent\textbf{Trajectory Prediction Results}. As shown in \Cref{fig: e2e_vis_append}, our model demonstrates strong trajectory planning in various challenging scenarios. It effectively handles decisions such as yielding to vehicles and pedestrians, navigating hazards in adverse weather, and accounting for occluded objects in low-visibility conditions. These examples highlight the model's robustness and adaptability, even in long-tail cases.

\section{Limitations}
\noindent\textbf{Limitations \& Future Work.}
We plan to explore \textit{more efficient and higher-accuracy architectures} for multi-task joint training (perception \& trajectory prediction), such as Mixture-of-Experts (MoE), where different tasks will be routed to specialized experts. In addition, for both perception and end-to-end trajectory prediction, we aim to employ \textit{a unified reinforcement learning framework with a multi-objective reward design} to jointly optimize these tasks.

%% file: main.bbl
\begin{thebibliography}{100}

\bibitem{hwang2024emma}
Jyh-Jing Hwang, Runsheng Xu, Hubert Lin, Wei-Chih Hung, Jingwei Ji, Kristy Choi, Di~Huang, Tong He, Paul Covington, Benjamin Sapp, et~al.
\newblock Emma: End-to-end multimodal model for autonomous driving.
\newblock {\em arXiv preprint arXiv:2410.23262}, 2024.

\bibitem{zheng2025diffusion}
Yinan Zheng, Ruiming Liang, Kexin Zheng, Jinliang Zheng, Liyuan Mao, Jianxiong Li, Weihao Gu, Rui Ai, Shengbo~Eben Li, Xianyuan Zhan, et~al.
\newblock Diffusion-based planning for autonomous driving with flexible guidance.
\newblock {\em arXiv preprint arXiv:2501.15564}, 2025.

\bibitem{fan2023autonomous}
Rui Fan, Sicen Guo, and Mohammud~Junaid Bocus.
\newblock Autonomous driving perception.
\newblock {\em Cham, Switzerland: Springer}, 2023.

\bibitem{tian2025nuscenes}
Kexin Tian, Jingrui Mao, Yunlong Zhang, Jiwan Jiang, Yang Zhou, and Zhengzhong Tu.
\newblock Nuscenes-spatialqa: A spatial understanding and reasoning benchmark for vision-language models in autonomous driving.
\newblock {\em arXiv preprint arXiv:2504.03164}, 2025.

\bibitem{guo2024drivemllm}
Xianda Guo, Ruijun Zhang, Yiqun Duan, Yuhang He, Chenming Zhang, Shuai Liu, and Long Chen.
\newblock Drivemllm: A benchmark for spatial understanding with multimodal large language models in autonomous driving.
\newblock {\em arXiv e-prints}, pages arXiv--2411, 2024.

\bibitem{wang2025alpamayo}
Yan Wang, Wenjie Luo, Junjie Bai, Yulong Cao, Tong Che, Ke~Chen, Yuxiao Chen, Jenna Diamond, Yifan Ding, Wenhao Ding, et~al.
\newblock Alpamayo-r1: Bridging reasoning and action prediction for generalizable autonomous driving in the long tail.
\newblock {\em arXiv preprint arXiv:2511.00088}, 2025.

\bibitem{fu2025orion}
Haoyu Fu, Diankun Zhang, Zongchuang Zhao, Jianfeng Cui, Dingkang Liang, Chong Zhang, Dingyuan Zhang, Hongwei Xie, Bing Wang, and Xiang Bai.
\newblock Orion: A holistic end-to-end autonomous driving framework by vision-language instructed action generation.
\newblock {\em arXiv preprint arXiv:2503.19755}, 2025.

\bibitem{xu2024vlm}
Yi~Xu, Yuxin Hu, Zaiwei Zhang, Gregory~P Meyer, Siva~Karthik Mustikovela, Siddhartha Srinivasa, Eric~M Wolff, and Xin Huang.
\newblock Vlm-ad: End-to-end autonomous driving through vision-language model supervision.
\newblock {\em arXiv preprint arXiv:2412.14446}, 2024.

\bibitem{wen2025dexvla}
Junjie Wen, Yichen Zhu, Jinming Li, Zhibin Tang, Chaomin Shen, and Feifei Feng.
\newblock Dexvla: Vision-language model with plug-in diffusion expert for general robot control.
\newblock {\em arXiv preprint arXiv:2502.05855}, 2025.

\bibitem{wang2025rad}
Yujin Wang, Quanfeng Liu, Zhengxin Jiang, Tianyi Wang, Junfeng Jiao, Hongqing Chu, Bingzhao Gao, and Hong Chen.
\newblock Rad: Retrieval-augmented decision-making of meta-actions with vision-language models in autonomous driving.
\newblock In {\em Proceedings of the Computer Vision and Pattern Recognition Conference}, pages 3838--3848, 2025.

\bibitem{pan2025omnimanip}
Mingjie Pan, Jiyao Zhang, Tianshu Wu, Yinghao Zhao, Wenlong Gao, and Hao Dong.
\newblock Omnimanip: Towards general robotic manipulation via object-centric interaction primitives as spatial constraints.
\newblock In {\em Proceedings of the Computer Vision and Pattern Recognition Conference}, pages 17359--17369, 2025.

\bibitem{argus2025cvla}
Max Argus, Jelena Bratulic, Houman Masnavi, Maxim Velikanov, Nick Heppert, Abhinav Valada, and Thomas Brox.
\newblock cvla: Towards efficient camera-space vlas.
\newblock {\em arXiv preprint arXiv:2507.02190}, 2025.

\bibitem{yang2025thinking}
Jihan Yang, Shusheng Yang, Anjali~W Gupta, Rilyn Han, Li~Fei-Fei, and Saining Xie.
\newblock Thinking in space: How multimodal large language models see, remember, and recall spaces.
\newblock In {\em Proceedings of the Computer Vision and Pattern Recognition Conference}, pages 10632--10643, 2025.

\bibitem{daxberger2025mm}
Erik Daxberger, Nina Wenzel, David Griffiths, Haiming Gang, Justin Lazarow, Gefen Kohavi, Kai Kang, Marcin Eichner, Yinfei Yang, Afshin Dehghan, et~al.
\newblock Mm-spatial: Exploring 3d spatial understanding in multimodal llms.
\newblock In {\em Proceedings of the IEEE/CVF International Conference on Computer Vision}, pages 7395--7408, 2025.

\bibitem{cheng2024spatialrgpt}
An-Chieh Cheng, Hongxu Yin, Yang Fu, Qiushan Guo, Ruihan Yang, Jan Kautz, Xiaolong Wang, and Sifei Liu.
\newblock Spatialrgpt: Grounded spatial reasoning in vision-language models.
\newblock {\em Advances in Neural Information Processing Systems}, 37:135062--135093, 2024.

\bibitem{jia2024bench2drive}
Xiaosong Jia, Zhenjie Yang, Qifeng Li, Zhiyuan Zhang, and Junchi Yan.
\newblock Bench2drive: Towards multi-ability benchmarking of closed-loop end-to-end autonomous driving.
\newblock {\em Advances in Neural Information Processing Systems}, 37:819--844, 2024.

\bibitem{dauner2024navsim}
Daniel Dauner, Marcel Hallgarten, Tianyu Li, Xinshuo Weng, Zhiyu Huang, Zetong Yang, Hongyang Li, Igor Gilitschenski, Boris Ivanovic, Marco Pavone, et~al.
\newblock Navsim: Data-driven non-reactive autonomous vehicle simulation and benchmarking.
\newblock {\em Advances in Neural Information Processing Systems}, 37:28706--28719, 2024.

\bibitem{caesar2020nuscenes}
Holger Caesar, Varun Bankiti, Alex~H Lang, Sourabh Vora, Venice~Erin Liong, Qiang Xu, Anush Krishnan, Yu~Pan, Giancarlo Baldan, and Oscar Beijbom.
\newblock nuscenes: A multimodal dataset for autonomous driving.
\newblock In {\em Proceedings of the IEEE/CVF conference on computer vision and pattern recognition}, pages 11621--11631, 2020.

\bibitem{zeng2025futuresightdrive}
Shuang Zeng, Xinyuan Chang, Mengwei Xie, Xinran Liu, Yifan Bai, Zheng Pan, Mu~Xu, and Xing Wei.
\newblock Futuresightdrive: Thinking visually with spatio-temporal cot for autonomous driving.
\newblock {\em arXiv preprint arXiv:2505.17685}, 2025.

\bibitem{chi2025impromptu}
Haohan Chi, Huan-ang Gao, Ziming Liu, Jianing Liu, Chenyu Liu, Jinwei Li, Kaisen Yang, Yangcheng Yu, Zeda Wang, Wenyi Li, et~al.
\newblock Impromptu vla: Open weights and open data for driving vision-language-action models.
\newblock {\em arXiv preprint arXiv:2505.23757}, 2025.

\bibitem{wang2024drivecot}
Tianqi Wang, Enze Xie, Ruihang Chu, Zhenguo Li, and Ping Luo.
\newblock Drivecot: Integrating chain-of-thought reasoning with end-to-end driving.
\newblock {\em arXiv preprint arXiv:2403.16996}, 2024.

\bibitem{wang2023visionllm}
Wenhai Wang, Zhe Chen, Xiaokang Chen, Jiannan Wu, Xizhou Zhu, Gang Zeng, Ping Luo, Tong Lu, Jie Zhou, Yu~Qiao, et~al.
\newblock Visionllm: Large language model is also an open-ended decoder for vision-centric tasks.
\newblock {\em Advances in Neural Information Processing Systems}, 36:61501--61513, 2023.

\bibitem{driess2023palm}
Danny Driess, Fei Xia, Mehdi~SM Sajjadi, Corey Lynch, Aakanksha Chowdhery, Brian Ichter, Ayzaan Wahid, Jonathan Tompson, Quan Vuong, Tianhe Yu, et~al.
\newblock Palm-e: an embodied multimodal language model.
\newblock In {\em Proceedings of the 40th International Conference on Machine Learning}, pages 8469--8488, 2023.

\bibitem{liao2025diffusiondrive}
Bencheng Liao, Shaoyu Chen, Haoran Yin, Bo~Jiang, Cheng Wang, Sixu Yan, Xinbang Zhang, Xiangyu Li, Ying Zhang, Qian Zhang, et~al.
\newblock Diffusiondrive: Truncated diffusion model for end-to-end autonomous driving.
\newblock In {\em Proceedings of the Computer Vision and Pattern Recognition Conference}, pages 12037--12047, 2025.

\bibitem{zhao2025diffe2e}
Rui Zhao, Yuze Fan, Ziguo Chen, Fei Gao, and Zhenhai Gao.
\newblock Diffe2e: Rethinking end-to-end driving with a hybrid action diffusion and supervised policy.
\newblock {\em arXiv preprint arXiv:2505.19516}, 2025.

\bibitem{liu2025bridgedrive}
Shu Liu, Wenlin Chen, Weihao Li, Zheng Wang, Lijin Yang, Jianing Huang, Yipin Zhang, Zhongzhan Huang, Ze~Cheng, and Hao Yang.
\newblock Bridgedrive: Diffusion bridge policy for closed-loop trajectory planning in autonomous driving.
\newblock {\em arXiv preprint arXiv:2509.23589}, 2025.

\bibitem{mao2023gpt}
Jiageng Mao, Yuxi Qian, Junjie Ye, Hang Zhao, and Yue Wang.
\newblock Gpt-driver: Learning to drive with gpt.
\newblock {\em arXiv preprint arXiv:2310.01415}, 2023.

\bibitem{zhang2024think}
Qiming Zhang, Meixin Zhu, and Hao~Frank Yang.
\newblock Think-driver: From driving-scene understanding to decision-making with vision language models.
\newblock In {\em European Conference on Computer Vision Workshop}, 2024.

\bibitem{hu2023planning}
Yihan Hu, Jiazhi Yang, Li~Chen, Keyu Li, Chonghao Sima, Xizhou Zhu, Siqi Chai, Senyao Du, Tianwei Lin, Wenhai Wang, et~al.
\newblock Planning-oriented autonomous driving.
\newblock In {\em Proceedings of the IEEE/CVF conference on computer vision and pattern recognition}, pages 17853--17862, 2023.

\bibitem{lin2014microsoft}
Tsung-Yi Lin, Michael Maire, Serge Belongie, James Hays, Pietro Perona, Deva Ramanan, Piotr Doll{\'a}r, and C~Lawrence Zitnick.
\newblock Microsoft coco: Common objects in context.
\newblock In {\em European conference on computer vision}, pages 740--755. Springer, 2014.

\bibitem{li2025lmm}
Jincheng Li, Chunyu Xie, Ji~Ao, Dawei Leng, and Yuhui Yin.
\newblock Lmm-det: Make large multimodal models excel in object detection.
\newblock In {\em Proceedings of the IEEE/CVF International Conference on Computer Vision}, pages 308--318, 2025.

\bibitem{lang2019pointpillars}
Alex~H Lang, Sourabh Vora, Holger Caesar, Lubing Zhou, Jiong Yang, and Oscar Beijbom.
\newblock Pointpillars: Fast encoders for object detection from point clouds.
\newblock In {\em Proceedings of the IEEE/CVF conference on computer vision and pattern recognition}, pages 12697--12705, 2019.

\bibitem{radford2021learning}
Alec Radford, Jong~Wook Kim, Chris Hallacy, Aditya Ramesh, Gabriel Goh, Sandhini Agarwal, Girish Sastry, Amanda Askell, Pamela Mishkin, Jack Clark, et~al.
\newblock Learning transferable visual models from natural language supervision.
\newblock In {\em International conference on machine learning}, pages 8748--8763. PmLR, 2021.

\bibitem{alayrac2022flamingo}
Jean-Baptiste Alayrac, Jeff Donahue, Pauline Luc, Antoine Miech, Iain Barr, Yana Hasson, Karel Lenc, Arthur Mensch, Katherine Millican, Malcolm Reynolds, et~al.
\newblock Flamingo: a visual language model for few-shot learning.
\newblock {\em Advances in neural information processing systems}, 35:23716--23736, 2022.

\bibitem{li2023blip}
Junnan Li, Dongxu Li, Silvio Savarese, and Steven Hoi.
\newblock Blip-2: Bootstrapping language-image pre-training with frozen image encoders and large language models.
\newblock In {\em International conference on machine learning}, pages 19730--19742. PMLR, 2023.

\bibitem{liu2023visual}
Haotian Liu, Chunyuan Li, Qingyang Wu, and Yong~Jae Lee.
\newblock Visual instruction tuning.
\newblock {\em Advances in neural information processing systems}, 36:34892--34916, 2023.

\bibitem{stogiannidis2025mind}
Ilias Stogiannidis, Steven McDonagh, and Sotirios~A Tsaftaris.
\newblock Mind the gap: Benchmarking spatial reasoning in vision-language models.
\newblock {\em arXiv preprint arXiv:2503.19707}, 2025.

\bibitem{brohan2024rt}
Anthony Brohan, Noah Brown, Justice Carbajal, Yevgen Chebotar, Xi~Chen, Krzysztof Choromanski, Tianli Ding, Danny Driess, Avinava Dubey, Chelsea Finn, et~al.
\newblock Rt-2: Vision-language-action models transfer web knowledge to robotic control, 2023.
\newblock {\em URL https://arxiv. org/abs/2307.15818}, 2024.

\bibitem{kim2024openvla}
Moo~Jin Kim, Karl Pertsch, Siddharth Karamcheti, Ted Xiao, Ashwin Balakrishna, Suraj Nair, Rafael Rafailov, Ethan Foster, Grace Lam, Pannag Sanketi, et~al.
\newblock Openvla: An open-source vision-language-action model.
\newblock {\em arXiv preprint arXiv:2406.09246}, 2024.

\bibitem{tian2024drivevlm}
Xiaoyu Tian, Junru Gu, Bailin Li, Yicheng Liu, Yang Wang, Zhiyong Zhao, Kun Zhan, Peng Jia, Xianpeng Lang, and Hang Zhao.
\newblock Drivevlm: The convergence of autonomous driving and large vision-language models.
\newblock {\em arXiv preprint arXiv:2402.12289}, 2024.

\bibitem{sima2024drivelm}
Chonghao Sima, Katrin Renz, Kashyap Chitta, Li~Chen, Hanxue Zhang, Chengen Xie, Jens Bei{\ss}wenger, Ping Luo, Andreas Geiger, and Hongyang Li.
\newblock Drivelm: Driving with graph visual question answering.
\newblock In {\em European conference on computer vision}, pages 256--274. Springer, 2024.

\bibitem{philion2020lift}
Jonah Philion and Sanja Fidler.
\newblock Lift, splat, shoot: Encoding images from arbitrary camera rigs by implicitly unprojecting to 3d.
\newblock In {\em European conference on computer vision}, pages 194--210. Springer, 2020.

\bibitem{li2023bevdepth}
Yinhao Li, Zheng Ge, Guanyi Yu, Jinrong Yang, Zengran Wang, Yukang Shi, Jianjian Sun, and Zeming Li.
\newblock Bevdepth: Acquisition of reliable depth for multi-view 3d object detection.
\newblock In {\em Proceedings of the AAAI conference on artificial intelligence}, volume~37, pages 1477--1485, 2023.

\bibitem{peng2023bevsegformer}
Lang Peng, Zhirong Chen, Zhangjie Fu, Pengpeng Liang, and Erkang Cheng.
\newblock Bevsegformer: Bird's eye view semantic segmentation from arbitrary camera rigs.
\newblock In {\em Proceedings of the IEEE/CVF Winter Conference on Applications of Computer Vision}, pages 5935--5943, 2023.

\bibitem{zhang2023occformer}
Yunpeng Zhang, Zheng Zhu, and Dalong Du.
\newblock Occformer: Dual-path transformer for vision-based 3d semantic occupancy prediction.
\newblock In {\em Proceedings of the IEEE/CVF International Conference on Computer Vision}, pages 9433--9443, 2023.

\bibitem{liu2022bevfusion}
Zhijian Liu, Haotian Tang, Alexander Amini, Xinyu Yang, Huizi Mao, Daniela Rus, and Song Han.
\newblock Bevfusion: Multi-task multi-sensor fusion with unified bird's-eye view representation.
\newblock {\em arXiv preprint arXiv:2205.13542}, 2022.

\bibitem{chitta2022transfuser}
Kashyap Chitta, Aditya Prakash, Bernhard Jaeger, Zehao Yu, Katrin Renz, and Andreas Geiger.
\newblock Transfuser: Imitation with transformer-based sensor fusion for autonomous driving.
\newblock {\em IEEE transactions on pattern analysis and machine intelligence}, 45(11):12878--12895, 2022.

\bibitem{chen2024end}
Li~Chen, Penghao Wu, Kashyap Chitta, Bernhard Jaeger, Andreas Geiger, and Hongyang Li.
\newblock End-to-end autonomous driving: Challenges and frontiers.
\newblock {\em IEEE Transactions on Pattern Analysis and Machine Intelligence}, 2024.

\bibitem{caesar2021nuplan}
Holger Caesar, Juraj Kabzan, Kok~Seang Tan, Whye~Kit Fong, Eric Wolff, Alex Lang, Luke Fletcher, Oscar Beijbom, and Sammy Omari.
\newblock nuplan: A closed-loop ml-based planning benchmark for autonomous vehicles.
\newblock {\em arXiv preprint arXiv:2106.11810}, 2021.

\bibitem{qian2024nuscenes}
Tianwen Qian, Jingjing Chen, Linhai Zhuo, Yang Jiao, and Yu-Gang Jiang.
\newblock Nuscenes-qa: A multi-modal visual question answering benchmark for autonomous driving scenario.
\newblock In {\em Proceedings of the AAAI Conference on Artificial Intelligence}, volume~38, pages 4542--4550, 2024.

\bibitem{li2024womd}
Yiheng Li, Cunxin Fan, Chongjian Ge, Zhihao Zhao, Chenran Li, Chenfeng Xu, Huaxiu Yao, Masayoshi Tomizuka, Bolei Zhou, Chen Tang, et~al.
\newblock Womd-reasoning: A large-scale dataset for interaction reasoning in driving.
\newblock {\em arXiv preprint arXiv:2407.04281}, 2024.

\bibitem{malla2023drama}
Srikanth Malla, Chiho Choi, Isht Dwivedi, Joon~Hee Choi, and Jiachen Li.
\newblock Drama: Joint risk localization and captioning in driving.
\newblock In {\em Proceedings of the IEEE/CVF winter conference on applications of computer vision}, pages 1043--1052, 2023.

\bibitem{dosovitskiy2017carla}
Alexey Dosovitskiy, German Ros, Felipe Codevilla, Antonio Lopez, and Vladlen Koltun.
\newblock Carla: An open urban driving simulator.
\newblock In {\em Conference on robot learning}, pages 1--16. PMLR, 2017.

\bibitem{nie2024reason2drive}
Ming Nie, Renyuan Peng, Chunwei Wang, Xinyue Cai, Jianhua Han, Hang Xu, and Li~Zhang.
\newblock Reason2drive: Towards interpretable and chain-based reasoning for autonomous driving.
\newblock In {\em European Conference on Computer Vision}, pages 292--308. Springer, 2024.

\bibitem{chen2024expanding}
Zhe Chen, Weiyun Wang, Yue Cao, Yangzhou Liu, Zhangwei Gao, Erfei Cui, Jinguo Zhu, Shenglong Ye, Hao Tian, Zhaoyang Liu, et~al.
\newblock Expanding performance boundaries of open-source multimodal models with model, data, and test-time scaling.
\newblock {\em arXiv preprint arXiv:2412.05271}, 2024.

\bibitem{tang2025ufo}
Hao Tang, Chenwei Xie, Haiyang Wang, Xiaoyi Bao, Tingyu Weng, Pandeng Li, Yun Zheng, and Liwei Wang.
\newblock Ufo: A unified approach to fine-grained visual perception via open-ended language interface.
\newblock {\em arXiv preprint arXiv:2503.01342}, 2025.

\bibitem{vaswani2017attention}
Ashish Vaswani, Noam Shazeer, Niki Parmar, Jakob Uszkoreit, Llion Jones, Aidan~N Gomez, {\L}ukasz Kaiser, and Illia Polosukhin.
\newblock Attention is all you need.
\newblock {\em Advances in neural information processing systems}, 30, 2017.

\bibitem{chen2021pix2seq}
Ting Chen, Saurabh Saxena, Lala Li, David~J Fleet, and Geoffrey Hinton.
\newblock Pix2seq: A language modeling framework for object detection.
\newblock {\em arXiv preprint arXiv:2109.10852}, 2021.

\bibitem{li2022grounded}
Liunian~Harold Li, Pengchuan Zhang, Haotian Zhang, Jianwei Yang, Chunyuan Li, Yiwu Zhong, Lijuan Wang, Lu~Yuan, Lei Zhang, Jenq-Neng Hwang, et~al.
\newblock Grounded language-image pre-training.
\newblock In {\em Proceedings of the IEEE/CVF conference on computer vision and pattern recognition}, pages 10965--10975, 2022.

\bibitem{minderer2022simple}
Matthias Minderer, Alexey Gritsenko, Austin Stone, Maxim Neumann, Dirk Weissenborn, Alexey Dosovitskiy, Aravindh Mahendran, Anurag Arnab, Mostafa Dehghani, Zhuoran Shen, et~al.
\newblock Simple open-vocabulary object detection.
\newblock In {\em European conference on computer vision}, pages 728--755. Springer, 2022.

\bibitem{chen2024far}
Zhe Chen, Weiyun Wang, Hao Tian, Shenglong Ye, Zhangwei Gao, Erfei Cui, Wenwen Tong, Kongzhi Hu, Jiapeng Luo, Zheng Ma, et~al.
\newblock How far are we to gpt-4v? closing the gap to commercial multimodal models with open-source suites.
\newblock {\em Science China Information Sciences}, 67(12):220101, 2024.

\bibitem{jiang2025detect}
Qing Jiang, Junan Huo, Xingyu Chen, Yuda Xiong, Zhaoyang Zeng, Yihao Chen, Tianhe Ren, Junzhi Yu, and Lei Zhang.
\newblock Detect anything via next point prediction.
\newblock {\em arXiv preprint arXiv:2510.12798}, 2025.

\bibitem{zhang2024calibrating}
Mozhi Zhang, Mianqiu Huang, Rundong Shi, Linsen Guo, Chong Peng, Peng Yan, Yaqian Zhou, and Xipeng Qiu.
\newblock Calibrating the confidence of large language models by eliciting fidelity.
\newblock {\em arXiv preprint arXiv:2404.02655}, 2024.

\bibitem{jiang2018acquisition}
Borui Jiang, Ruixuan Luo, Jiayuan Mao, Tete Xiao, and Yuning Jiang.
\newblock Acquisition of localization confidence for accurate object detection.
\newblock In {\em Proceedings of the European conference on computer vision (ECCV)}, pages 784--799, 2018.

\bibitem{li2020generalized}
Xiang Li, Wenhai Wang, Lijun Wu, Shuo Chen, Xiaolin Hu, Jun Li, Jinhui Tang, and Jian Yang.
\newblock Generalized focal loss: Learning qualified and distributed bounding boxes for dense object detection.
\newblock {\em Advances in neural information processing systems}, 33:21002--21012, 2020.

\bibitem{lin2017focal}
Tsung-Yi Lin, Priya Goyal, Ross Girshick, Kaiming He, and Piotr Doll{\'a}r.
\newblock Focal loss for dense object detection.
\newblock In {\em Proceedings of the IEEE international conference on computer vision}, pages 2980--2988, 2017.

\bibitem{milletari2016v}
Fausto Milletari, Nassir Navab, and Seyed-Ahmad Ahmadi.
\newblock V-net: Fully convolutional neural networks for volumetric medical image segmentation.
\newblock In {\em 2016 fourth international conference on 3D vision (3DV)}, pages 565--571. Ieee, 2016.

\bibitem{li2024bevformer}
Zhiqi Li, Wenhai Wang, Hongyang Li, Enze Xie, Chonghao Sima, Tong Lu, Qiao Yu, and Jifeng Dai.
\newblock Bevformer: learning bird's-eye-view representation from lidar-camera via spatiotemporal transformers.
\newblock {\em IEEE Transactions on Pattern Analysis and Machine Intelligence}, 2024.

\bibitem{huang2021bevdet}
Junjie Huang, Guan Huang, Zheng Zhu, Yun Ye, and Dalong Du.
\newblock Bevdet: High-performance multi-camera 3d object detection in bird-eye-view.
\newblock {\em arXiv preprint arXiv:2112.11790}, 2021.

\bibitem{shi2016real}
Wenzhe Shi, Jose Caballero, Ferenc Husz{\'a}r, Johannes Totz, Andrew~P Aitken, Rob Bishop, Daniel Rueckert, and Zehan Wang.
\newblock Real-time single image and video super-resolution using an efficient sub-pixel convolutional neural network.
\newblock In {\em Proceedings of the IEEE conference on computer vision and pattern recognition}, pages 1874--1883, 2016.

\bibitem{he2017mask}
Kaiming He, Georgia Gkioxari, Piotr Doll{\'a}r, and Ross Girshick.
\newblock Mask r-cnn.
\newblock In {\em Proceedings of the IEEE international conference on computer vision}, pages 2961--2969, 2017.

\bibitem{chen2022unified}
Ting Chen, Saurabh Saxena, Lala Li, Tsung-Yi Lin, David~J Fleet, and Geoffrey~E Hinton.
\newblock A unified sequence interface for vision tasks.
\newblock {\em Advances in Neural Information Processing Systems}, 35:31333--31346, 2022.

\bibitem{wang2024git}
Haiyang Wang, Hao Tang, Li~Jiang, Shaoshuai Shi, Muhammad~Ferjad Naeem, Hongsheng Li, Bernt Schiele, and Liwei Wang.
\newblock Git: Towards generalist vision transformer through universal language interface.
\newblock In {\em European Conference on Computer Vision}, pages 55--73. Springer, 2024.

\bibitem{zhang2022dino}
Hao Zhang, Feng Li, Shilong Liu, Lei Zhang, Hang Su, Jun Zhu, Lionel~M Ni, and Heung-Yeung Shum.
\newblock Dino: Detr with improved denoising anchor boxes for end-to-end object detection.
\newblock {\em arXiv preprint arXiv:2203.03605}, 2022.

\bibitem{wang2021fcos3d}
Tai Wang, Xinge Zhu, Jiangmiao Pang, and Dahua Lin.
\newblock Fcos3d: Fully convolutional one-stage monocular 3d object detection.
\newblock In {\em Proceedings of the IEEE/CVF international conference on computer vision}, pages 913--922, 2021.

\bibitem{cheng2022masked}
Bowen Cheng, Ishan Misra, Alexander~G Schwing, Alexander Kirillov, and Rohit Girdhar.
\newblock Masked-attention mask transformer for universal image segmentation.
\newblock In {\em Proceedings of the IEEE/CVF conference on computer vision and pattern recognition}, pages 1290--1299, 2022.

\bibitem{caesar2018coco}
Holger Caesar, Jasper Uijlings, and Vittorio Ferrari.
\newblock Coco-stuff: Thing and stuff classes in context.
\newblock In {\em Proceedings of the IEEE conference on computer vision and pattern recognition}, pages 1209--1218, 2018.

\bibitem{zhan2024griffon}
Yufei Zhan, Hongyin Zhao, Yousong Zhu, Fan Yang, Ming Tang, and Jinqiao Wang.
\newblock Griffon-g: Bridging vision-language and vision-centric tasks via large multimodal models.
\newblock {\em arXiv preprint arXiv:2410.16163}, 2024.

\bibitem{ma2024groma}
Chuofan Ma, Yi~Jiang, Jiannan Wu, Zehuan Yuan, and Xiaojuan Qi.
\newblock Groma: Localized visual tokenization for grounding multimodal large language models.
\newblock In {\em European Conference on Computer Vision}, pages 417--435. Springer, 2024.

\bibitem{liu2025vlm}
Peng Liu, Haozhan Shen, Chunxin Fang, Zhicheng Sun, Jiajia Liao, and Tiancheng Zhao.
\newblock Vlm-fo1: Bridging the gap between high-level reasoning and fine-grained perception in vlms.
\newblock {\em arXiv preprint arXiv:2509.25916}, 2025.

\bibitem{girshick2015fast}
Ross Girshick.
\newblock Fast r-cnn.
\newblock In {\em Proceedings of the IEEE international conference on computer vision}, pages 1440--1448, 2015.

\bibitem{sun2020scalability}
Pei Sun, Henrik Kretzschmar, Xerxes Dotiwalla, Aurelien Chouard, Vijaysai Patnaik, Paul Tsui, James Guo, Yin Zhou, Yuning Chai, Benjamin Caine, et~al.
\newblock Scalability in perception for autonomous driving: Waymo open dataset.
\newblock In {\em Proceedings of the IEEE/CVF conference on computer vision and pattern recognition}, pages 2446--2454, 2020.

\bibitem{zhou2019semantic}
Bolei Zhou, Hang Zhao, Xavier Puig, Tete Xiao, Sanja Fidler, Adela Barriuso, and Antonio Torralba.
\newblock Semantic understanding of scenes through the ade20k dataset.
\newblock {\em International Journal of Computer Vision}, 127(3):302--321, 2019.

\bibitem{kazemzadeh2014referitgame}
Sahar Kazemzadeh, Vicente Ordonez, Mark Matten, and Tamara Berg.
\newblock Referitgame: Referring to objects in photographs of natural scenes.
\newblock In {\em Proceedings of the 2014 conference on empirical methods in natural language processing (EMNLP)}, pages 787--798, 2014.

\bibitem{yu2016modeling}
Licheng Yu, Patrick Poirson, Shan Yang, Alexander~C Berg, and Tamara~L Berg.
\newblock Modeling context in referring expressions.
\newblock In {\em European conference on computer vision}, pages 69--85. Springer, 2016.

\bibitem{mao2016generation}
Junhua Mao, Jonathan Huang, Alexander Toshev, Oana Camburu, Alan~L Yuille, and Kevin Murphy.
\newblock Generation and comprehension of unambiguous object descriptions.
\newblock In {\em Proceedings of the IEEE conference on computer vision and pattern recognition}, pages 11--20, 2016.

\bibitem{marcu2023lingoqa}
Ana-Maria Marcu, Long Chen, Jan Hünermann, Alice Karnsund, Benoit Hanotte, Prajwal Chidananda, Saurabh Nair, Vijay Badrinarayanan, Alex Kendall, Jamie Shotton, and Oleg Sinavski.
\newblock Lingoqa: Visual question answering for autonomous driving.
\newblock {\em arXiv preprint arXiv:2312.14115}, 2023.

\bibitem{li2024automated}
Yanze Li, Wenhua Zhang, Kai Chen, Yanxin Liu, Pengxiang Li, Ruiyuan Gao, Lanqing Hong, Meng Tian, Xinhai Zhao, Zhenguo Li, et~al.
\newblock Automated evaluation of large vision-language models on self-driving corner cases.
\newblock {\em arXiv preprint arXiv:2404.10595}, 2024.

\bibitem{ma2024dolphins}
Yingzi Ma, Yulong Cao, Jiachen Sun, Marco Pavone, and Chaowei Xiao.
\newblock Dolphins: Multimodal language model for driving.
\newblock In {\em European Conference on Computer Vision}, pages 403--420. Springer, 2024.

\bibitem{lu2025can}
Yuhang Lu, Yichen Yao, Jiadong Tu, Jiangnan Shao, Yuexin Ma, and Xinge Zhu.
\newblock Can lvlms obtain a driver’s license? a benchmark towards reliable agi for autonomous driving.
\newblock In {\em Proceedings of the AAAI Conference on Artificial Intelligence}, volume~39, pages 5838--5846, 2025.

\bibitem{cao2024maplm}
Xu~Cao, Tong Zhou, Yunsheng Ma, Wenqian Ye, Can Cui, Kun Tang, Zhipeng Cao, Kaizhao Liang, Ziran Wang, James~M Rehg, et~al.
\newblock Maplm: A real-world large-scale vision-language benchmark for map and traffic scene understanding.
\newblock In {\em Proceedings of the IEEE/CVF conference on computer vision and pattern recognition}, pages 21819--21830, 2024.

\bibitem{xu2024drivegpt4}
Zhenhua Xu, Yujia Zhang, Enze Xie, Zhen Zhao, Yong Guo, Kwan-Yee~K Wong, Zhenguo Li, and Hengshuang Zhao.
\newblock Drivegpt4: Interpretable end-to-end autonomous driving via large language model.
\newblock {\em IEEE Robotics and Automation Letters}, 2024.

\bibitem{jiang2023vad}
Bo~Jiang, Shaoyu Chen, Qing Xu, Bencheng Liao, Jiajie Chen, Helong Zhou, Qian Zhang, Wenyu Liu, Chang Huang, and Xinggang Wang.
\newblock Vad: Vectorized scene representation for efficient autonomous driving.
\newblock In {\em Proceedings of the IEEE/CVF International Conference on Computer Vision}, pages 8340--8350, 2023.

\bibitem{chen2024vadv2}
Shaoyu Chen, Bo~Jiang, Hao Gao, Bencheng Liao, Qing Xu, Qian Zhang, Chang Huang, Wenyu Liu, and Xinggang Wang.
\newblock Vadv2: End-to-end vectorized autonomous driving via probabilistic planning.
\newblock {\em arXiv preprint arXiv:2402.13243}, 2024.

\bibitem{li2024ego}
Zhiqi Li, Zhiding Yu, Shiyi Lan, Jiahan Li, Jan Kautz, Tong Lu, and Jose~M Alvarez.
\newblock Is ego status all you need for open-loop end-to-end autonomous driving?
\newblock In {\em Proceedings of the IEEE/CVF Conference on Computer Vision and Pattern Recognition}, pages 14864--14873, 2024.

\bibitem{yuan2024drama}
Chengran Yuan, Zhanqi Zhang, Jiawei Sun, Shuo Sun, Zefan Huang, Christina Dao~Wen Lee, Dongen Li, Yuhang Han, Anthony Wong, Keng~Peng Tee, et~al.
\newblock Drama: An efficient end-to-end motion planner for autonomous driving with mamba.
\newblock {\em arXiv preprint arXiv:2408.03601}, 2024.

\bibitem{li2024hydra}
Zhenxin Li, Kailin Li, Shihao Wang, Shiyi Lan, Zhiding Yu, Yishen Ji, Zhiqi Li, Ziyue Zhu, Jan Kautz, Zuxuan Wu, et~al.
\newblock Hydra-mdp: End-to-end multimodal planning with multi-target hydra-distillation.
\newblock {\em arXiv preprint arXiv:2406.06978}, 2024.

\bibitem{loshchilov2017decoupled}
Ilya Loshchilov and Frank Hutter.
\newblock Decoupled weight decay regularization.
\newblock {\em arXiv preprint arXiv:1711.05101}, 2017.

\bibitem{micikevicius2017mixed}
Paulius Micikevicius, Sharan Narang, Jonah Alben, Gregory Diamos, Erich Elsen, David Garcia, Boris Ginsburg, Michael Houston, Oleksii Kuchaiev, Ganesh Venkatesh, et~al.
\newblock Mixed precision training.
\newblock {\em arXiv preprint arXiv:1710.03740}, 2017.

\bibitem{chen2016training}
Tianqi Chen, Bing Xu, Chiyuan Zhang, and Carlos Guestrin.
\newblock Training deep nets with sublinear memory cost.
\newblock {\em arXiv preprint arXiv:1604.06174}, 2016.

\bibitem{yan2018second}
Yan Yan, Yuxing Mao, and Bo~Li.
\newblock Second: Sparsely embedded convolutional detection.
\newblock {\em Sensors}, 18(10):3337, 2018.

\bibitem{zhou2025autovla}
Zewei Zhou, Tianhui Cai, Seth~Z Zhao, Yun Zhang, Zhiyu Huang, Bolei Zhou, and Jiaqi Ma.
\newblock Autovla: A vision-language-action model for end-to-end autonomous driving with adaptive reasoning and reinforcement fine-tuning.
\newblock {\em arXiv preprint arXiv:2506.13757}, 2025.

\bibitem{xu2025streamingvlm}
Ruyi Xu, Guangxuan Xiao, Yukang Chen, Liuning He, Kelly Peng, Yao Lu, and Song Han.
\newblock Streamingvlm: Real-time understanding for infinite video streams.
\newblock {\em arXiv preprint arXiv:2510.09608}, 2025.

\bibitem{ning2025livevlm}
Zhenyu Ning, Guangda Liu, Qihao Jin, Wenchao Ding, Minyi Guo, and Jieru Zhao.
\newblock Livevlm: Efficient online video understanding via streaming-oriented kv cache and retrieval.
\newblock {\em arXiv preprint arXiv:2505.15269}, 2025.

\bibitem{xiao2023efficient}
Guangxuan Xiao, Yuandong Tian, Beidi Chen, Song Han, and Mike Lewis.
\newblock Efficient streaming language models with attention sinks.
\newblock {\em arXiv preprint arXiv:2309.17453}, 2023.

\bibitem{yao2025cacheblend}
Jiayi Yao, Hanchen Li, Yuhan Liu, Siddhant Ray, Yihua Cheng, Qizheng Zhang, Kuntai Du, Shan Lu, and Junchen Jiang.
\newblock Cacheblend: Fast large language model serving for rag with cached knowledge fusion.
\newblock In {\em Proceedings of the Twentieth European Conference on Computer Systems}, pages 94--109, 2025.

\bibitem{witte2025epipolar}
Christian Witte, Jens Behley, Cyrill Stachniss, and Marvin Raaijmakers.
\newblock Epipolar attention field transformers for bird's eye view semantic segmentation.
\newblock In {\em 2025 IEEE/CVF Winter Conference on Applications of Computer Vision (WACV)}, pages 8660--8669. IEEE, 2025.

\bibitem{li2025generalized}
Zhenxin Li, Wenhao Yao, Zi~Wang, Xinglong Sun, Joshua Chen, Nadine Chang, Maying Shen, Zuxuan Wu, Shiyi Lan, and Jose~M Alvarez.
\newblock Generalized trajectory scoring for end-to-end multimodal planning.
\newblock {\em arXiv preprint arXiv:2506.06664}, 2025.

\bibitem{kamath2021mdetr}
Aishwarya Kamath, Mannat Singh, Yann LeCun, Gabriel Synnaeve, Ishan Misra, and Nicolas Carion.
\newblock Mdetr-modulated detection for end-to-end multi-modal understanding.
\newblock In {\em Proceedings of the IEEE/CVF international conference on computer vision}, pages 1780--1790, 2021.

\bibitem{liu2024grounding}
Shilong Liu, Zhaoyang Zeng, Tianhe Ren, Feng Li, Hao Zhang, Jie Yang, Qing Jiang, Chunyuan Li, Jianwei Yang, Hang Su, et~al.
\newblock Grounding dino: Marrying dino with grounded pre-training for open-set object detection.
\newblock In {\em European conference on computer vision}, pages 38--55. Springer, 2024.

\bibitem{chen2023shikra}
Keqin Chen, Zhao Zhang, Weili Zeng, Richong Zhang, Feng Zhu, and Rui Zhao.
\newblock Shikra: Unleashing multimodal llm's referential dialogue magic.
\newblock {\em arXiv preprint arXiv:2306.15195}, 2023.

\bibitem{chen2023minigpt}
Jun Chen, Deyao Zhu, Xiaoqian Shen, Xiang Li, Zechun Liu, Pengchuan Zhang, Raghuraman Krishnamoorthi, Vikas Chandra, Yunyang Xiong, and Mohamed Elhoseiny.
\newblock Minigpt-v2: large language model as a unified interface for vision-language multi-task learning.
\newblock {\em arXiv preprint arXiv:2310.09478}, 2023.

\bibitem{pramanick2024jack}
Shraman Pramanick, Guangxing Han, Rui Hou, Sayan Nag, Ser-Nam Lim, Nicolas Ballas, Qifan Wang, Rama Chellappa, and Amjad Almahairi.
\newblock Jack of all tasks master of many: Designing general-purpose coarse-to-fine vision-language model.
\newblock In {\em Proceedings of the IEEE/CVF Conference on Computer Vision and Pattern Recognition}, pages 14076--14088, 2024.

\bibitem{wu2024general}
Junfeng Wu, Yi~Jiang, Qihao Liu, Zehuan Yuan, Xiang Bai, and Song Bai.
\newblock General object foundation model for images and videos at scale.
\newblock In {\em Proceedings of the IEEE/CVF Conference on Computer Vision and Pattern Recognition}, pages 3783--3795, 2024.

\bibitem{yan2023universal}
Bin Yan, Yi~Jiang, Jiannan Wu, Dong Wang, Ping Luo, Zehuan Yuan, and Huchuan Lu.
\newblock Universal instance perception as object discovery and retrieval.
\newblock In {\em Proceedings of the IEEE/CVF Conference on Computer Vision and Pattern Recognition}, pages 15325--15336, 2023.

\bibitem{lai2024lisa}
Xin Lai, Zhuotao Tian, Yukang Chen, Yanwei Li, Yuhui Yuan, Shu Liu, and Jiaya Jia.
\newblock Lisa: Reasoning segmentation via large language model.
\newblock In {\em Proceedings of the IEEE/CVF Conference on Computer Vision and Pattern Recognition}, pages 9579--9589, 2024.

\bibitem{rasheed2024glamm}
Hanoona Rasheed, Muhammad Maaz, Sahal Shaji, Abdelrahman Shaker, Salman Khan, Hisham Cholakkal, Rao~M Anwer, Eric Xing, Ming-Hsuan Yang, and Fahad~S Khan.
\newblock Glamm: Pixel grounding large multimodal model.
\newblock In {\em Proceedings of the IEEE/CVF Conference on Computer Vision and Pattern Recognition}, pages 13009--13018, 2024.

\bibitem{wang2025himtok}
Tao Wang, Changxu Cheng, Lingfeng Wang, Senda Chen, and Wuyue Zhao.
\newblock Himtok: Learning hierarchical mask tokens for image segmentation with large multimodal model.
\newblock {\em arXiv preprint arXiv:2503.13026}, 2025.

\end{thebibliography}
